\documentclass[10pt,twocolumn,letterpaper]{article}

\usepackage[applications]{wacv}

\usepackage{multirow}

\newcommand{\ourmethod}{\textsc{Gestalt}\xspace}
\newcommand{\geneflow}{GeneFlow\xspace}
\newcommand{\pairbiask}{\ensuremath{k_{\text{pair}}}\xspace}
\newcommand{\unicos}{\ensuremath{\text{embCOS}}\xspace}

\makeatletter
\def\@secpenalty{0}

\def\@afterheading{%
\@nobreaktrue
  \everypar{%
    \if@nobreak
      \@nobreakfalse
      \clubpenalty 300\relax
      \if@afterindent \else {\setbox\z@\lastbox}\fi
    \else
      \clubpenalty \@clubpenalty
      \everypar{}%
    \fi}}
\makeatother

\AtEndPreamble{
  \crefname{table}{Table}{Tables}
  \Crefname{table}{Table}{Tables}
  \crefname{figure}{Figure}{Figures}
  \Crefname{figure}{Figure}{Figures}
}

\definecolor{wacvblue}{rgb}{0.21,0.49,0.74}
\usepackage[pagebackref,breaklinks,colorlinks,allcolors=wacvblue]{hyperref}

\usepackage[accsupp]{axessibility}

\def\wacvPaperID{723}
\def\confName{WACV}
\def\confYear{2027}

\title{GESTALT: Fast, Generalizable Histopathology Image Synthesis from \\
       Gene Expression with Pretrained Pathology Foundation Models}

\author{Felix Ho\\
Stony Brook University\\
Stony Brook, NY\\
{\tt\small nhan.ho@stonybrook.edu}
\and
Alexandros Graikos\\
Stony Brook University\\
Stony Brook, NY\\
{\tt\small agraikos@cs.stonybrook.edu}
\and
Dimitris Samaras\\
Stony Brook University\\
Stony Brook, NY\\
{\tt\small samaras@cs.stonybrook.edu}
}

\begin{document}
\maketitle
\begin{abstract}
Spatial transcriptomics measures gene expression while preserving where each
cell sits in the tissue. The inverse, what a given transcriptomic state
looks like, is less explored but equally useful: it lets researchers
visualize the morphological consequence of a molecular state and interpret a
spatial assay's molecular calls against real tissue structure.
\geneflow, the current state of the art on the Xenium gene expression-to-image benchmark, trains a model from scratch, incurring
a high cost on every new target dataset it is applied to. We instead propose
\ourmethod, which reformulates gene expression-to-image synthesis as predicting
a pathology foundation model's image embedding. A lightweight
gene-token transformer predicts this embedding from multi-cell gene
expression, and a frozen or LoRA-adapted diffusion decoder (PixCell) renders
it as an image. Because this visual knowledge is reused rather than relearned,
\ourmethod trains in a fraction of the time. In both multi-cell and single-cell settings, our method
outperforms the SoTA on perceptual metrics across all three Xenium datasets.
Beyond improved image quality, \ourmethod natively synthesizes at
$1024\times1024$, and is evaluated in three generalization settings: zero-shot transfer
across gene panels and datasets, held-out patients and disease states, and
transfer across studies and tissue sources. Coupled with a
statistical model of spatial expression, \ourmethod also
provides, to our knowledge, the first paired
synthesis of cell-resolved transcriptomic fields and tissue images with
explicit control over cell type, density, and composition.
Code is available at \url{https://github.com/trongnhandev1011/Gestalt}.
\end{abstract}

\section{Introduction}
\label{sec:intro}

Spatial transcriptomics platforms such as 10x Xenium and Visium
measure gene expression while preserving the spatial context of the tissue~\cite{chen2023spatial,tian2023expanding}.
A substantial body of literature aims to address the prediction of expression
\emph{from} an H\&E image~\cite{spirit,ghist,stem}. The opposite direction,
generating the tissue image that corresponds to a given transcriptomic
state, has received considerably less attention despite comparable
utility~\cite{geneflow}.
Cancer is a natural setting in which to study it, since molecular
alterations manifest as diverse and clinically consequential histological
patterns.

The current state of the art in Xenium expression-to-image synthesis is
\geneflow~\cite{geneflow}, which jointly trains an RNA encoder and a
conditional rectified flow \cite{liu2023flow} generative model for each dataset.
Adapting it to a new gene panel therefore requires retraining both
the conditioning and generation pipelines, without reusing any components
that may share information across datasets.

At the same time, pathology foundation models (FMs)~\cite{uni,uni2,conch,virchow2}
can produce general-purpose representations of H\&E tissue that can be used for
both discriminative and, more recently, generative tasks.
Generative models have been trained to synthesize tissue
conditioned on foundation model H\&E embeddings~\cite{pixcell, duboudin2026cytosyn, xiang2026generative},
and taken together, a foundation and a generative model provide
an image representation and a renderer that
can be reused rather than learned anew for each target task.

Therefore, instead of training a gene expression-to-image generative
model from scratch, we propose to formulate expression-to-image synthesis as a two-stage process:
we first learn to map gene expressions to pathology foundation model image embeddings,
a supervised regression problem that is decoupled from generating tissue images.
Once this mapping is learned, we can utilize a pre-trained decoder that renders tissue tiles
from such embeddings, which can be used with little to no adaptation.

Following this formulation, we introduce \ourmethod
(\textbf{G}ene \textbf{E}xpression \textbf{S}ynthesis of \textbf{T}issue
\textbf{A}ppearance via \textbf{L}atent \textbf{T}ransfer), a lightweight
gene-token transformer for predicting pathology FM tile embeddings from
cell-resolved spatial gene expression, paired with a pre-trained generative model to synthesize
H\&E tiles from such embeddings. \ourmethod exceeds \geneflow in terms of generated
image quality on its benchmark datasets (\cref{sec:exp-indist}), while also adapting
to each target dataset in a fraction of the time (\cref{sec:exp-cost}).

Generalization is critical for a gene expression-to-image model, since
spatial transcriptomics cohorts vary in gene panel, patient, disease state,
study, and institution. We test whether the learned molecular-to-morphological
mapping survives these shifts through zero-shot transfer across gene panels and
datasets, held-out patients and disease states, and transfer across studies and
tissue sources. We also test whether expression is a meaningful control space by
varying cell type, density, and composition in statistically generated
expression fields, enabling targeted paired-data synthesis without treating
uncertain image-derived expression as ground truth~\cite{i2gbench}.

Our contributions can be summarized as:
\begin{itemize}
\item \textbf{A new formulation for gene expression-to-image synthesis.}
      We generate images from gene expression by predicting a pathology FM's
      image embedding. A lightweight gene-token transformer predicts this
      embedding from the cells in a tile, and a pretrained decoder turns it
      into an image.
\item \textbf{Improvement over the SoTA at a fraction of the cost.}
      By reusing the FM and decoder, we outperform the current
      state-of-the-art on three Xenium panels, improving FID by $21$--$32$ points,
      while also requiring substantially fewer trainable parameters and
      training time.
\item \textbf{Generalization and scaling.}
      We showcase zero-shot transfer across gene panels and datasets, studies, and tissue sources, and
      generalization to held-out patients and disease states.
      Additionally, we can easily swap the FM or decoder
      to change the tile embedding space we operate in,
      or enable $1024{\times}1024$ gene-to-image synthesis.
\item \textbf{Controllable paired synthesis.} Coupled with a statistical
      model of spatial expression, \ourmethod provides, to our knowledge, the
      first paired synthesis of cell-resolved transcriptomic fields and tissue
      images, with explicit control over cell type, density, and composition.
\end{itemize}

\section{Related work}
\label{sec:related}

\paragraph{Pathology foundation models.} Large-scale self-supervised
pretraining has produced histology FMs such as UNI~\cite{uni},
UNI2~\cite{uni2}, CONCH~\cite{conch}, and Virchow2~\cite{virchow2}. Their tile
embeddings transfer across tasks and organs with little or no fine-tuning.
Recent work has further leveraged pretrained pathology representations
for weakly supervised segmentation through structural distillation
\cite{le2026construct}, while disentangled stain and morphological
representations have been explored for robust histopathology analysis
under limited annotations~\cite{pham2026disentangled}.
Diffusion models have been used to extend this capability to generation,
learning to render a realistic tile from a foundation model embedding~\cite{graikos2024learned,yellapragada2025zoomldm}.
PixCell~\cite{pixcell} scales up the diffusion training to cover multiple cancer types,
and learns to decode UNI2-h embeddings~\cite{uni2} into H\&E tissue images.
\ourmethod learns the missing map from spatial gene expression into this
`renderable' embedding space, allowing the pretrained image representation and
decoder to be reused.

\paragraph{Expression-to-image generation.} \geneflow~\cite{geneflow} is our
closest directly comparable method, mapping single- or multi-cell expression
to paired histology with an attention-based RNA encoder and a conditional
UNet trained by rectified flow. Since its encoder and generator are trained
jointly per dataset, adaptation requires retraining the generative model.
\ourmethod instead separates a lightweight gene-to-embedding mapper from a
pretrained decoder, reused directly or adapted with LoRA.
GE2Hist~\cite{ge2hist} also generates local histology from single-cell
expression, but conditions on one central cell whose profile is derived
from Visium HD. ST2HE~\cite{st2he} generates virtual H\&E from
$512{\times}512$ tiles of DAPI morphology overlaid with transcript locations,
rather than expression alone. Concurrent methods use other inputs.
STMDiT~\cite{vlachas2026transcriptomics}, also built on PixCell, conditions
on a slide-level UNI2-h embedding together with a Visium spot
profile. SPATIA~\cite{kong2025spatia} uses a control cell image and a
transition descriptor. MuPD~\cite{xiang2026generative} conditions on bulk-RNA
pathway scores rather than cell-resolved expression. Our input is the measured expression and coordinates of every
Xenium cell in the field; differences in platform, conditioning, and evaluation protocol prevent a matched comparison.

\paragraph{Cross-modal prediction and alignment.} Image-to-expression methods
include SPiRiT~\cite{spirit}, which uses a vision transformer over
Xenium-matched tiles; GHIST~\cite{ghist}, which incorporates cell,
neighborhood, and nuclear cues; and STEM~\cite{stem}, which models uncertainty with
conditional diffusion. They address the reverse direction and are not directly
comparable to ours.

Other methods align transcriptomics and histology in a shared space.
OmiCLIP~\cite{omiclip} aligns gene descriptions with images for retrieval,
STORM~\cite{storm} learns through masked multimodal reconstruction, and
H\&Enium~\cite{henium} projects pathology and transcriptomic FM embeddings.
None of them generate images from the aligned embedding. We instead render
predicted embeddings and evaluate the resulting images.

\section{Method}
\label{sec:method}

\begin{figure*}[t]
\centering
\includegraphics[width=1.0\linewidth]{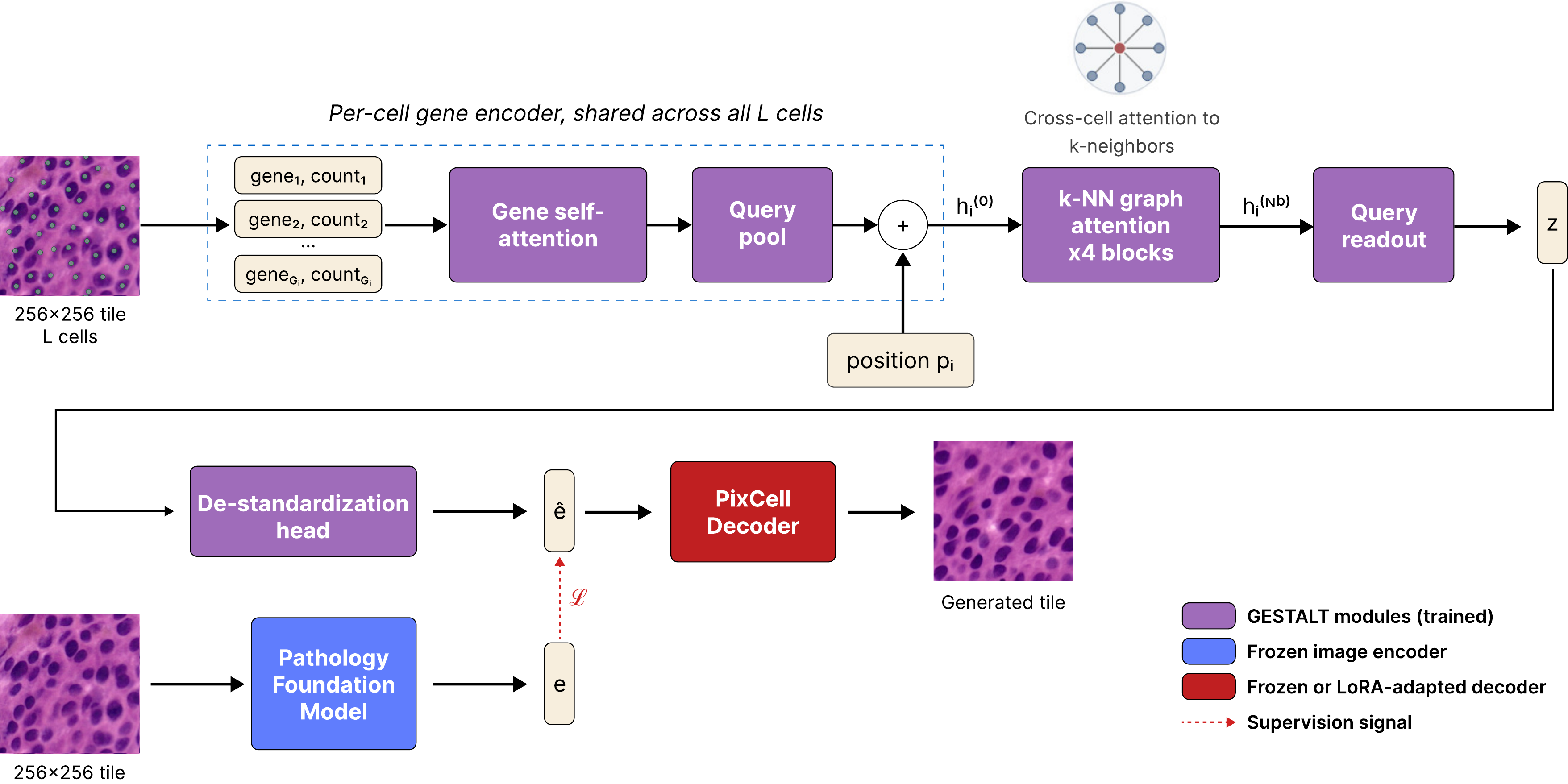}
\caption{\textbf{\ourmethod.} Each cell is encoded along the gene axis and
pooled into one token, cells are mixed only with their $k{=}8$ nearest
neighbors, and a query readout produces the tile vector $z$. A
de-standardization head maps $z$ into a frozen FM's embedding space, which
PixCell renders. Only the purple components are trained; training regresses
$\hat e$ onto the FM's embedding of the same tile (\cref{eq:loss})}
\label{fig:pipeline}
\end{figure*}

\subsection{Gene expressions to image embeddings}
\label{sec:method-backbone}
Given a spatial expression field, \ourmethod first predicts the embedding
that a pathology FM assigns to the corresponding tissue tile. A pretrained
decoder then renders the predicted embedding into an H\&E image. The mapper
preserves the hierarchical structure of the input by encoding genes within
each cell before modeling interactions across nearby cells.
\cref{fig:pipeline} summarizes the complete pipeline.

\paragraph{Notation.} A tile comprises $L$ segmented cells. Each cell $i$ is
given by its centroid $(x_i, y_i)$ and its sparse expression
$c_i = \{(g_{ij}, v_{ij})\}_{j=1}^{G_i}$, where $g_{ij}$ indexes a gene in
the panel's vocabulary and $v_{ij}$ is that gene's standardized count. The
mapper $f_\theta$ is trained to map the set $\{(x_i, y_i, c_i)\}_{i=1}^{L}$ to a
single tile image embedding $\hat e$. We present it as a composition of three stages,
$f_\theta = \operatorname{DeStd} \left( f_\text{cell} \left( f_\text{gene} \left( \{ c_1, \dots, c_L \} \right) \right) \right)$: a
per-cell gene encoder, a cross-cell $k$-NN attention stack, and a
de-standardization head, each discussed below.
We use hidden width $H{=}384$ and $N_b{=}4$ cross-cell blocks.

\paragraph{Per-cell gene encoding.} A gene and its count form one token,
$t_{ij} = \operatorname{LN}\!\big(E[g_{ij}] + \phi(v_{ij})\big) \in \mathbb{R}^{H}$,
with $E$ a learned gene embedding and $\phi$ a small MLP on the count.
These tokens self-attend along the gene axis \emph{within} cell $i$ only,
and a learned query $q$ pools them into one cell token, giving the first
stage
\begin{equation}
    \ell_i = f_{\text{gene}}(c_i) = \operatorname{Pool}_q\!\big(\operatorname{Transformer}_\text{gene}(\{t_{ij}\}_{j=1}^{G_i})\big),
    \label{eq:cellenc}
\end{equation}
with $\ell_i \in \mathbb{R}^{H}$. $\operatorname{Pool}_q$ is single-query
attention rather than a mean, allowing each gene's contribution to be learned
rather than fixed at $1/G_i$. Pooling is within a cell, leaving $L$ cell
tokens for the spatial stage below.

\paragraph{Cross-cell attention over a sparse $k$-NN graph.} A linear
position embedding $p_i = W_p [x_i, y_i]^\top + b_p$ is added to each cell
token, giving $h_i^{(0)} = \ell_i + p_i$. Cells then
exchange information through $N_b$ stacked pre-norm
graph-attention (GAT) blocks
after which a single learned
query pools the refined cell tokens into the tile token, giving the second
stage $f_\text{cell}$:
\begin{equation}
\begin{aligned}
    h^{(n)} &= \operatorname{GAT}^{(n)}_{k\text{-NN}}\!\big(h^{(n-1)}\big),
              \quad n = 1,\dots,N_b,\\[2pt]
    z       &= \operatorname{Pool}_\text{tile}\!\big(h^{(N_b)}\big),
\end{aligned}
\label{eq:backbone}
\end{equation}
where $z \in \mathbb{R}^{H}$ is the pooled
tile token.
Each cell attends only to
itself and a $k$-nearest-neighbor set built from cell coordinates once per
forward pass,
\begin{equation}
    \mathcal{N}_k(i) = \operatorname*{arg\,topk}_{j \neq i}\; -\lVert [x_i,y_i]^T - [x_j, y_j]^T \rVert_2 .
    \label{eq:knn}
\end{equation}
Restricting attention to the nearest $k$ focuses each cell's summary on its
local microenvironment.
The attention logit between $i$ and neighbor $j$ receives
a learned pairwise bias on their relative offset,
\begin{equation}
    s_{ij} = \frac{q_i \cdot k_j}{\sqrt{d_h}} + \mathrm{MLP}_\text{edge}\big([\Delta x_{ij}, \Delta y_{ij}, \lVert \Delta_{ij}\rVert]\big),
\end{equation}
where $d_h = H / n_\text{heads} = 48$ is the per-head width ($n_\text{heads}{=}8$), so closeness and
direction modulate a neighbor's contribution even among the $k$ retained
neighbors. $k$ (\pairbiask) is the
single hyperparameter controlling each cell's effective receptive field; we
use $k=8$ for all main experiments.

\paragraph{Predicting an image embedding.}
\label{sec:method-mapper}
A de-standardization head maps the pooled tile token $z$ of
\cref{eq:backbone} to the target embedding domain and undoes the standardization
applied to the training targets,
\begin{equation}
    \hat e
           = \operatorname{DeStd}(z) = (W z' + b) \odot \sigma + \mu \in \mathbb{R}^{d},
    \label{eq:destd}
\end{equation}
with $z' = \operatorname{LN}(z)$ and $(\mu, \sigma)$ the per-dimension mean
and standard deviation of the training targets. $\hat e$ is the embedding a
frozen pathology FM would assign to the tile's real image.

\paragraph{Training.}
The mapping function $f_\theta$ is
trained end-to-end by directly regressing real image embeddings as
\begin{equation}
    \mathcal{L} = \frac{1}{d}\sum_{j=1}^{d}
        \Big(\frac{\hat e_j - e_j}{\sigma_j}\Big)^{\!2}
      + \lambda_\text{c}\big(1 - \cos(\hat e, e)\big)
      + \lambda_\text{n}\big\lvert \lVert \hat e \rVert - \lVert e \rVert \big\rvert,
    \label{eq:loss}
\end{equation}
with $\lambda_\text{c} = 0.25$ and $\lambda_\text{n} = 0.1$. The first term
is a squared error standardized per dimension by the same $\sigma$ as
\cref{eq:destd}, so no coordinate dominates by scale; the second constrains
the direction of $\hat e$; the third keeps its norm from drifting away from
the target's. Predicting an embedding rather than pixels inherits PixCell's
decoder architecture, resolution and sampling algorithm for free:
$f_\theta$ only has to land in a space PixCell already renders.
Unlike prior work that contrastively aligns expression and image modalities
(OmiCLIP~\cite{omiclip}, STORM~\cite{storm},
H\&Enium~\cite{henium}), we never train for cross-modal alignment;
the mapper projects gene expressions into UNI2-h's space by direct supervision alone. A frozen or
LoRA-adapted decoder $D$ then renders $\hat I = D(\hat e)$, so with $D$
fixed, the modeling burden falls on $f_\theta$.

\subsection{Decoder adaptation}
PixCell \cite{pixcell} is a latent diffusion transformer conditioned on a
foundation-model image embedding. We adapt it with LoRA \cite{hu2022lora},
applying adapters in all $28$ transformer blocks to the QKV and output
projections of self- and cross-attention and to both feed-forward matrices.
This includes the cross-attention through which the conditioning enters.
We use rank $r{=}32$ and $\alpha{=}64$, with zero initialization so that
adaptation starts from the pretrained decoder. This adds $26.8$M trainable
parameters, $3.9\%$ of PixCell's $691.9$M parameters. The adapters are
optimized for 8000 steps using PixCell's diffusion denoising objective
\cite{pixcell}, separately from the gene-to-embedding projection.

Training the mapper and decoder adapter separately makes their errors
attributable. Alongside image metrics, we report \unicos between $\hat e$ and
the real $e$, and oracle image metrics obtained by rendering $D(e)$. The
oracle-to-ground-truth gap measures renderer error, while the mapper-to-oracle
gap measures the effect of replacing $e$ with $\hat e$. We report both because
they need not move together. \Cref{sec:supp-lora} expands on this comparison.

\subsection{\texorpdfstring{Scaling to $1024{\times}1024$ resolution}%
                           {Scaling to 1024x1024 resolution}}
\label{sec:method-1024}
The base PixCell model synthesizes tiles at $256\times256$ resolution.
We can also utilize PixCell-1024, which is conditioned on a grid of $4\times4$ embeddings, one per
$256$px sub-region. \ourmethod is used to predict all sixteen embeddings
in a single forward pass over the field, and neither the gene encoder nor
the graph-attention stack changes. A field is presented as one set of
cells whose coordinates are normalized to $[-1,1]$ over the full $1024$px extent, so
the $k$-NN graph of \cref{eq:knn} spans the union of the sixteen
sub-regions: a cell near a  sub-region boundary
attends to neighbors on the other side of it, and spatial context crosses
boundaries before any sub-region is read out.

The readout is the only component that changes. The single pooling query
of \cref{eq:backbone} is replaced by sixteen queries $\{u_m\}_{m=1}^{16}$
with learned content and \emph{fixed} positions $r_m$ at the sub-region
centers,
\begin{equation}
\begin{aligned}
    s_{mi} &= \frac{u_m \cdot W_k h_i}{\sqrt{d_h}}
              + \mathrm{MLP}'_\text{edge}\big([\Delta_{mi}, \lVert \Delta_{mi} \rVert]\big), \\[2pt]
    z_m    &= \textstyle\sum_{i=1}^{L} \operatorname{softmax}_i(s_{mi})\, W_v h_i,
\end{aligned}
\label{eq:gridreadout}
\end{equation}
where $h_i = h_i^{(N_b)}$, $\Delta_{mi} = r_m - [x_i, y_i]^T$,
and $\mathrm{MLP}'_\text{edge}$ is an edge network of the same form as the
backbone's. The head of \cref{eq:destd} is then applied to each $z_m$,
giving $\smash{\hat E} = [\hat e_1, \dots, \hat e_{16}]$.

\paragraph{Keeping each query local.} Each query should describe its own
sub-region, so that every part of the field is rendered from the cells beneath
it. The learned bias alone puts only $38$--$51\%$ of a query's attention there,
blending neighboring sub-regions together, so we penalize attention to distant
cells,
\begin{equation}
    s_{mi} \;\leftarrow\; s_{mi} - \gamma \lVert \Delta_{mi} \rVert^2 ,
    \label{eq:readoutdecay}
\end{equation}
with $\Delta_{mi}$ in normalized units (one unit is $512$px) and $\gamma$ a
learnable per-head weight initialized at $50$. Each query then draws
$88$--$92\%$ from its own sub-region, while the backbone still mixes cells
across sub-region boundaries.

\section{Experiments}
\label{sec:experiments}

\subsection{Setup}
\textbf{Metrics.} We report FID, SSIM, and FeatDist (paired Inception-2048
$\ell_2$ distance) following \geneflow~\cite{geneflow}. Two further diagnostics
help identify the errors introduced by our two-stage process:
\unicos, the cosine similarity between the predicted
$\hat{e}$ and real embedding $e$, and oracle FID, obtained by giving the same
decoder the real embedding. Unless otherwise stated, we use \pairbiask{=}8,
UNI2-h, and rank-32 LoRA. PixCell is pretrained on PanCan-30M~\cite{pixcell}.
Its documented sources are TCGA, CPTAC, GTEx, SBU, an ovarian cancer cohort,
and NLST, none of which appear among our Xenium and NCBI/GEO cohorts.

\subsection{Xenium C1, C2 and P1}
\label{sec:exp-genexen}

\paragraph{Datasets.} We use the three Xenium melanoma FFPE samples
\geneflow~\cite{geneflow} reports on, Xenium$_{C1}$ (282 genes, $100.6$k
cells), Xenium$_{C2}$ (382 genes, C1's panel nested inside, $78.7$k cells)
and Xenium$_{P1}$ (5006-gene Prime panel, $98.0$k cells), using its released data and code. We compare against \geneflow's published results and restrict
superiority claims to these benchmarks, since \geneflow does not report our
HEST protocols.

\paragraph{In-distribution.}
\label{sec:exp-indist}
Train and test tiles come from the same dataset, using a random 80/20 tile
split as in \geneflow's released code. Both sides are evaluated at the native $40\times$ (0.25\,$\mu$m/px),
and we retrain the mapper and its LoRA adapter at this resolution.

\begin{table}[tb]
\centering\small
\begin{tabular}{lcccc}
\toprule
 & FID $\downarrow$ & SSIM $\uparrow$ & FeatDist $\downarrow$ & \unicos $\uparrow$ \\
\midrule
\multicolumn{5}{l}{\emph{\geneflow~\cite{geneflow} (multi-cell, reported, $40\times$)}} \\
Xenium$_{C1}$ & 47.95 & 0.28 & 14.50 & -- \\
Xenium$_{C2}$ & 45.50 & \textbf{0.40} & 15.61 & -- \\
Xenium$_{P1}$ & 34.31 & 0.23 & 13.41 & -- \\
\midrule
\multicolumn{5}{l}{\emph{Ours (LoRA, $40\times$)}} \\
Xenium$_{C1}$ & \textbf{15.75} & \textbf{0.350} & \textbf{12.66} & 0.838 \\
Xenium$_{C2}$ & \textbf{18.03} & 0.396 & \textbf{12.47} & 0.860 \\
Xenium$_{P1}$ & \textbf{13.15} & \textbf{0.283} & \textbf{12.52} & 0.807 \\
\midrule
\multicolumn{5}{l}{\emph{Oracle (real embedding, $40\times$), decoder reference}} \\
Xenium$_{C1}$ & \emph{14.27} & \emph{0.353} & \emph{11.94} & -- \\
Xenium$_{C2}$ & \emph{14.75} & \emph{0.400} & \emph{11.69} & -- \\
Xenium$_{P1}$ & \emph{11.54} & \emph{0.287} & \emph{11.73} & -- \\
\bottomrule
\end{tabular}
\caption{In-distribution comparison against \geneflow's reported numbers at
its native $40\times$ (0.25\,$\mu$m/px). GESTALT and oracle rows report means
over seeds 7, 42 and 13. Standard deviations and native $20\times$ results
are in \cref{sec:supp-indist-full}.}
\label{tab:indist}
\end{table}

\begin{figure*}[tb]
\centering
\includegraphics[width=1.0\linewidth]{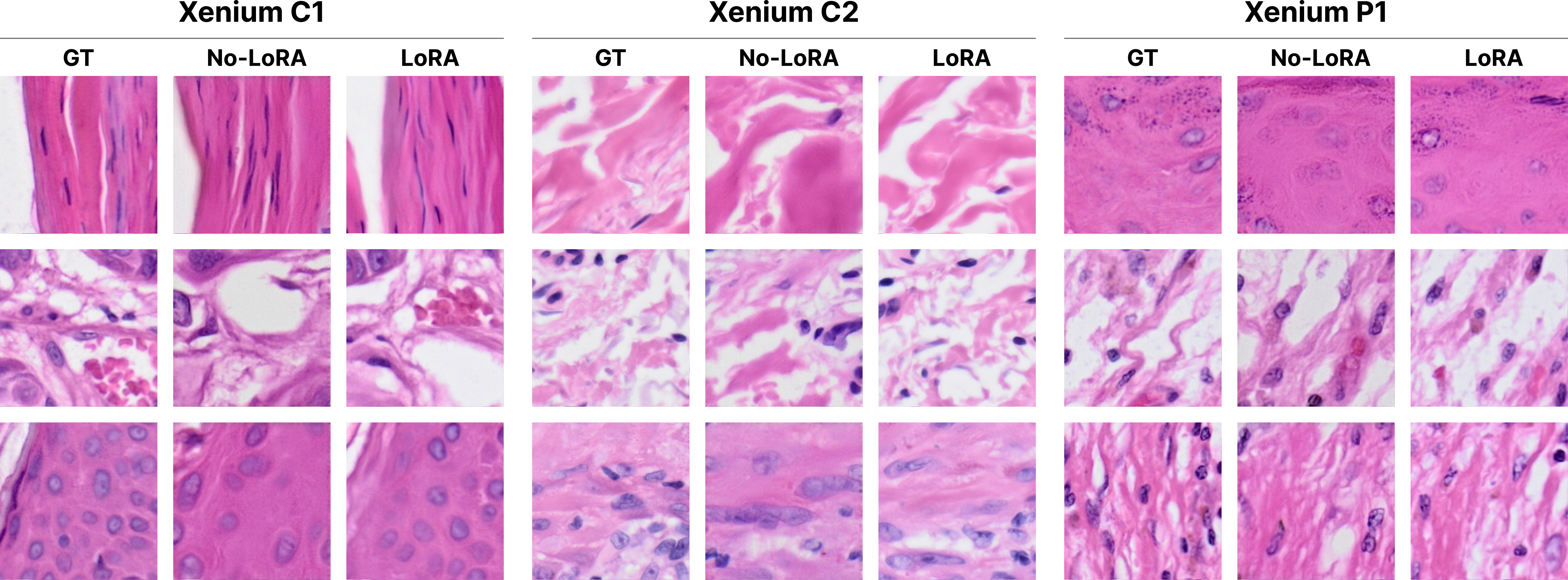}
\caption{Qualitative in-distribution comparison across Xenium C1, C2 and P1
at native $40\times$. Rows are held-out tiles spanning sparse to moderate
cell density: real tile (GT), frozen decoder (No-LoRA), adapted decoder (LoRA).}
\label{fig:qualitative-c1c2p1}
\end{figure*}

\Cref{tab:indist} shows that \ourmethod improves FID by $21$ to $32$ points across the three panels,
FeatDist on all three, and SSIM on two of the three. The mapper is
$1.5$ to $3.3$ FID above the oracle reference and predicts the real
embedding at a cosine of $0.81$ to $0.86$. The relatively small
prediction-to-oracle gaps indicate that much of the remaining error is
already present when the decoder receives the real embedding. On
Xenium$_{C2}$, the oracle SSIM is $0.400$, matching \geneflow's $0.40$ at
its reported precision, while our predicted-embedding result reaches
$0.396$. Qualitative comparisons across the three panels are shown in
\cref{fig:qualitative-c1c2p1}.

\geneflow further evaluates on single-cell conditioning, where a window is centered
on one cell and conditioned only on its expression. In \cref{sec:supp-indist-full}
we show that \ourmethod also improves perceptual metrics in the single-cell setting.

\paragraph{Cross-dataset generalization.}
\label{sec:exp-crossdata}
\geneflow also trains on one Xenium panel and evaluates zero-shot on another,
restricting both sides to the 126 genes the three panels share. We repeat that
test at \geneflow's native $40\times$, training on one of C1/C2/P1 and
evaluating on another, all six directions, with the mapper and its LoRA fitted
only on the source. \ourmethod attains a better FID on five of six directions, by more
than $50$ points on two, and FeatDist on four (\cref{tab:crossgeneflow}).

\begin{table}[tb]
\centering\small
\resizebox{\linewidth}{!}{%
\begin{tabular}{lccccc}
\toprule
\multirow{2}{*}{Direction} & \multicolumn{2}{c}{\geneflow} & \multicolumn{2}{c}{\ourmethod} & Oracle \\
\cmidrule(lr){2-3} \cmidrule(lr){4-5} \cmidrule(lr){6-6}
 & FID & FeatDist & FID & FeatDist & FID \\
\midrule
C1 $\to$ C2 & 112.25 & 17.18 & \textbf{62.71} & \textbf{16.23} & 15.24 \\
C2 $\to$ C1 & 146.78 & 18.36 & \textbf{138.65} & \textbf{18.08} & 25.65 \\
C2 $\to$ P1 & 148.15 & 17.59 & \textbf{89.18} & \textbf{16.47} & 19.54 \\
P1 $\to$ C2 & 127.79 & 17.27 & \textbf{75.24} & \textbf{15.62} & 15.04 \\
C1 $\to$ P1 & 79.86 & \textbf{15.80} & \textbf{56.62} & 16.53 & 11.52 \\
\midrule
P1 $\to$ C1 & \textbf{89.16} & \textbf{16.11} & 112.05 & 17.37 & 18.95 \\
\bottomrule
\end{tabular}
}
\caption{\textbf{Cross-dataset generalization}, all six directions across
C1, C2, P1, matching \geneflow's Table~4 protocol: the shared 126-gene
subset, native $40\times$ on both sides. Oracle FID varies with the source
as well as the target because the LoRA is fitted on the source.}
\label{tab:crossgeneflow}
\end{table}

\subsection{HEST Xenium}
\label{sec:exp-crosscond}

The preceding in-distribution benchmark holds out tiles while allowing the
same slide to appear in training. The cross-dataset benchmark holds out the
target C1/C2/P1 slide and tests transfer across gene panels, but does not
establish separation by patient, disease state, study, or institution. We
therefore construct slide-disjoint HEST-Xenium evaluations to test transfer
beyond the original benchmark.

We evaluate lung and breast from HEST-1k~\cite{hest}, resampled to $20\times$
(0.5\,$\mu$m/px). Lung results are summarized below. Cross-study breast
transfer attains FID $97$--$106$ against oracle FID $28$--$41$, with
$80$--$92\%$ top-$50$ gene-importance agreement across held-out slides;
full results appear in \cref{sec:supp-breast}.

\paragraph{Lung: disease-matched slide holdouts.}
\label{sec:exp-lung}
\label{sec:exp-ssl-gen}
We train on 13 diseased lung slides and evaluate on two held-out diseased
sections, NCBI882 and NCBI861 (\cref{tab:lung-holdout}). \ourmethod reaches
FIDs of $38.22$ and $59.37$, within $9.6$ and $14.3$ of the oracle, and SSIM
within $0.006$ of the oracle on both sections. A mapper trained only on other
slides thus transfers to held-out sections without any section-specific training,
with FID close to the oracle's.

\begin{table}[tb]
\centering\small
\setlength{\tabcolsep}{4pt}
\begin{tabular}{lccc}
\toprule
 & FID $\downarrow$ & SSIM $\uparrow$ & FeatDist $\downarrow$ \\
\midrule
\multicolumn{4}{l}{\emph{NCBI882 (held-out diseased section)}} \\
\ourmethod & 38.22 & 0.162 & 12.69 \\
\emph{Oracle} & 28.64 & 0.168 & 11.23 \\
\midrule
\multicolumn{4}{l}{\emph{NCBI861 (held-out diseased section)}} \\
\ourmethod & 59.37 & 0.176 & 13.07 \\
\emph{Oracle} & 45.09 & 0.180 & 11.62 \\
\bottomrule
\end{tabular}
\caption{\textbf{Disease-matched slide holdouts, HEST lung.} Trained on
13 diseased slides and evaluated on two held-out diseased sections,
NCBI882 and NCBI861, excluded from all training. Oracle rows decode the
real image embedding.}
\label{tab:lung-holdout}
\end{table}

\paragraph{Lung: cross-condition generalization.}
We further train on all 15 diseased slides and hold out two healthy patients,
NCBI884 and NCBI875, so that both patient identity and disease state are
unseen. Even in this harder setting, \ourmethod still transfers to both
patients (\cref{tab:supp-lung-crosscond}), although the transfer is
slide-specific: on NCBI884 the FID of $45.86$ stays within $15.8$ of the
oracle's $30.07$, while on NCBI875 the gap widens to $42.6$ ($89.65$ against
$47.08$).

\subsection{Magnification and resolution}
\label{sec:exp-scale}
The same pipeline supports decoding at different magnifications and field sizes:

\emph{$20\times$ magnification:} We train the corresponding mapper and
adapter at $20\times$ (0.5\,$\mu$m/px), PixCell's native magnification, and
report the results in \cref{sec:supp-indist-full}.

\begin{figure*}[tb]
\centering
\includegraphics[width=\linewidth]{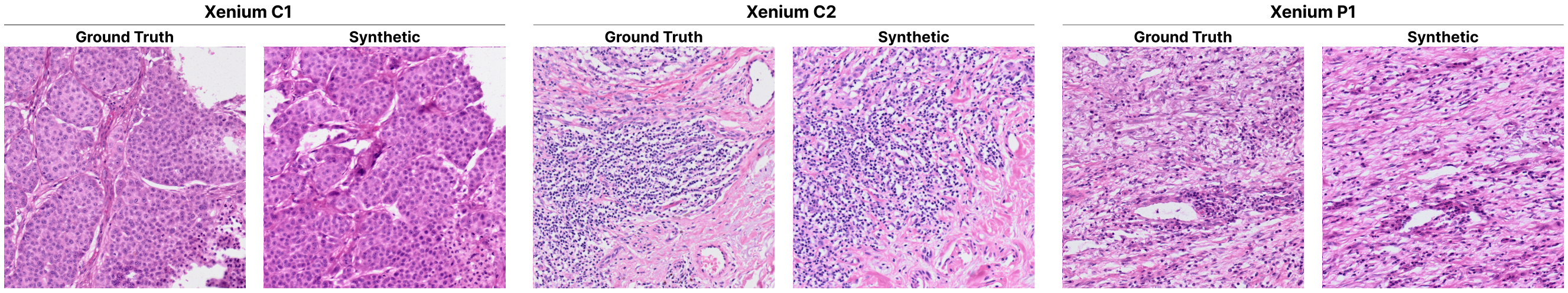}
\caption{\textbf{Full $1024\times1024$ region synthesis on Xenium C1, C2
and P1.} Each pair is a held-out field and its prediction, produced by a
single joint decoding pass over all sixteen $256\times256$ sub-regions with
no tiling seams or blending. Scores in \cref{tab:genehist1024}.}
\label{fig:qualitative-1024}
\end{figure*}

\emph{Larger-region synthesis at $1024{\times}1024$:} \geneflow, like most
gene-to-image methods, is trained and reported at $256{\times}256$. \ourmethod instead
synthesizes a field sixteen times larger through the grid readout of
\cref{sec:method-1024}, using a separately trained mapper. We report Crop FID
and CLIP FID~\cite{graikos2024learned,pixcell} in
\cref{tab:genehist1024}, with full-region examples in
\cref{fig:qualitative-1024}. To our knowledge, no prior expression-to-image
method operates at this resolution, so there is no direct baseline.
\Cref{sec:supp-grid1024} analyses the readout's locality and sweeps its decay.

\begin{table}[tb]
\centering\small
\begin{tabular}{lcccc}
\toprule
 & \multicolumn{2}{c}{Pred vs.\ GT} & \multicolumn{2}{c}{Oracle vs.\ GT} \\
 \cmidrule(lr){2-3} \cmidrule(lr){4-5}
 & Crop FID & CLIP FID & Crop FID & CLIP FID \\
\midrule
Xenium$_{C1}$ & 71.59 & 5.29 & 59.12 & 3.62 \\
Xenium$_{C2}$ & 50.87 & 3.33 & 39.20 & 2.08 \\
Xenium$_{P1}$ & 49.87 & 2.98 & 47.92 & 2.91 \\
\bottomrule
\end{tabular}
\caption{\textbf{\textsc{Gestalt}-1024} on C1/C2/P1, scored with Crop FID
and CLIP FID via Clean-FID~\cite{cleanfid}, the protocol PixCell reports for its
high-resolution model.}
\label{tab:genehist1024}
\end{table}

\subsection{Cost}
\label{sec:exp-cost}
\Cref{tab:cost} compares trainable parameters and wall-clock. In \geneflow's
released code, the encoder is rebuilt for each panel and adds $27$k parameters per gene,
primarily to predict both factors of a rank-$50$ gene-gene relation matrix.
For \ourmethod, only the gene-embedding table $E$ grows with the panel.

The parameter counts remain unchanged at $1024{\times}1024$, where mapper
training and decoder adaptation take $90$--$142$ and $34$ minutes,
respectively, or $124$--$176$ minutes in total. A $1024{\times}1024$ decode
takes $4.8$s against $1.7$s for one $256{\times}256$ tile, so it costs about
$5.7\times$ less per unit area.

Both stages ran on one 48GB GPU, peaking at 3.4GB for mapper training and
6.4GB for decoder adaptation. \geneflow reports ${\sim}12$h on an H100 and
up to 78GB at batch size 96, without a per-panel breakdown, so
\cref{tab:cost} repeats its time for C1, C2, and P1. On its released UNet, a
$640\times640$ training step costs $7.1\times$ more than at $256\times256$.
A fitted scaling model projects approximately $20\times$ the training time
at $1024\times1024$, or about ten days at its reported rate. This is an
extrapolation, not a direct training run.

\begin{table}[tb]
\centering\small
\setlength{\tabcolsep}{4pt}
\begin{tabular}{lcccc}
\toprule
& \multicolumn{3}{c}{Params (M)} & \\
\cmidrule(lr){2-4}
Method & $C1$ & $C2$ & $P1$ &
\shortstack{Wall-clock\\$256^2$ / $1024^2$} \\
\midrule
GeneFlow~\cite{geneflow}
    & 51.0 & 53.7 & 178.5
    & $\sim$12h / -- \\
\midrule
\textsc{Gestalt} & & & & \\[-0.2em]
Mapper
    & 7.7 & 7.7 & 9.5
    & 55--80 / 90--142 min \\
Decoder LoRA
    & 26.8 & 26.8 & 26.8
    & 34 / 34 min \\
\cmidrule(lr){1-5}
Total
    & 34.5 & 34.5 & 36.3
    & 89--114 / 124--176 min \\
\bottomrule
\end{tabular}
\caption{Adaptation cost per Xenium panel. GESTALT wall-clock is reported
as $256^2/1024^2$; parameter counts are unchanged across resolutions.
\geneflow's parameter counts are taken from its released code and its
${\sim}12$h time is self-reported, while ours are measured on one 48GB GPU.}
\label{tab:cost}
\end{table}

\section{Controllable tissue synthesis}
\label{sec:synthesis}

Jointly measuring histology and spatial expression is expensive. One
alternative is to infer expression from H\&E~\cite{stem,ghist,spirit}, but
current methods recover only part of the molecular signal. In a benchmark of
seven methods, the top-performing models predict most genes with average
Pearson correlations of $0.1$--$0.3$;
predictability increases with spatial autocorrelation and expression
magnitude and decreases with sparsity~\cite{i2gbench}.

An alternative is to generate expression first and then render its
corresponding tissue. Existing simulators such as scDesign3, scCube and
sCCIgen produce plausible spatial expression
fields~\cite{scdesign3,sccube,sccigen}, but not paired tissue images.
\ourmethod provides this
renderer, allowing synthetic cell-resolved fields to be paired with generated
tissue images. We demonstrate
this capability with a compact statistical expression model, although another
simulator could be used in its place.

\paragraph{A statistical expression model.} A synthetic field must represent
variation between fields, spatial correlation among nearby cells, and
realistic expression sparsity. We model these properties using a shared field
state, spatially correlated cell residuals, and a detected-gene-count model.

The generator works in two stages. After reducing cells of one type
to $100$ principal components, it represents each real tile by the mean of its
cells. For a synthetic field, it draws a shared tile state $s$ from a Gaussian
mixture fitted to these means, then draws each cell state $z_i$ around it:
\begin{gather}
 s \sim \sum_{m=1}^{M}\pi_m\mathcal{N}(\mu_m,\Sigma_m),\quad
 z_i=s+\epsilon_i,\\
 \operatorname{Cov}(\epsilon_i,\epsilon_j)
 =\exp\!\left(-\frac{\lVert r_i-r_j\rVert_2}{\rho}\right)\Sigma_\epsilon.
 \label{eq:expression-generator}
\end{gather}
Here $r_i$ is cell $i$'s position, $\Sigma_\epsilon$ is the residual
covariance in PCA space, and $\rho$ is a spatial range fitted from the data.
The number of mixture components $M$ is chosen over $1$--$20$ by
the Bayesian information criterion (BIC). This is a random-effects model with
an exponential covariance function~\cite{cressie1993spatial}. The number of
cells is drawn from the observed type-specific distribution, and their
positions follow a hard-core point process~\cite{diggle2013point}, whose
minimum separation adapts to that count. After mapping $z_i$ back to gene
space, the number of detected genes is regressed on the cell state to preserve
realistic sparsity.

The mixture components produce the different appearances. BIC selects $16$
components for C1 melanoma and $10$ for C1 keratinocytes, against $3$ for C2
keratinocytes and $2$ for C2 T cells. A single state reproduces a type's mean
expression but not the variation between its fields.

\begin{figure*}[!tb]
\centering
\includegraphics[width=1.0\linewidth]{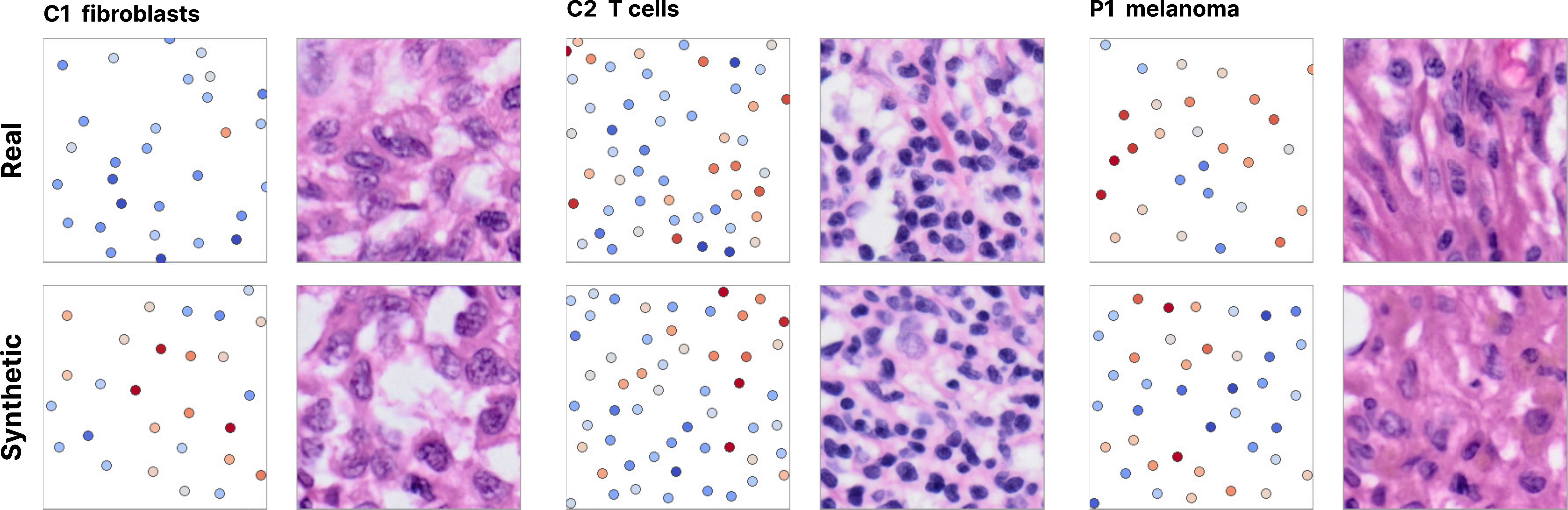}
\caption{\textbf{Synthesis by cell type.} Each expression map precedes its
image. The top row shows measured expression and H\&E from a held-out tile;
the bottom shows an independently drawn synthetic field of the same type and
its generated image. Color shows each cell's score on the type-specific
first principal component, using a scale shared across rows. Reference tiles
have the median cell count and field state for their type. The rows are not
paired, and no ground truth exists for the synthetic fields.}
\label{fig:synth-celltype}
\end{figure*}

\paragraph{Synthesis by cell type.} \Cref{fig:synth-celltype} shows a
synthetic field and its generated image for melanoma on P1, T cells on C2 and
fibroblasts on C1, each beside a real tile of the same type. Each field is
drawn from its type's expression model alone, with no measured expression or
image as input, and is rendered by the same mapper and decoder used
throughout.

\paragraph{Limitations.} Synthetic fields are less variable
than real tissue. On C1 melanoma, the mean pairwise cosine among $40$
predicted field embeddings is $0.82$, compared with $0.66$ among $40$ real
tiles of the same type. Replacing sampled states with those measured from real
tiles leaves this value unchanged, so the missing variation comes from
structure within each field, which the Gaussian model does not capture.
\Cref{sec:supp-synthesis} varies cell density and cell-type proportion;
generated CellViT nucleus counts preserve the requested density ordering.

\section{Ablations}
\label{sec:ablation}

Ablations on the in-distribution C1/C2/P1 splits are less informative because
train and test tiles may share slide-specific appearance, making weak variants
look competitive. We therefore perform the main ablations under the
cross-condition HEST lung setting of \cref{sec:exp-crosscond}, on the healthy
holdouts NCBI884 and NCBI875 (\cref{tab:supp-lung-crosscond}).

\paragraph{Does expression conditioning matter?}
Keeping the decoder fixed, we replace the predicted embedding with a constant,
a seeded permutation of the predictions, or the average real FM embedding of
the ten training tiles with the most similar expression. On both healthy holdouts,
\ourmethod outperforms the constant and permutation controls on both paired
metrics, and the retrieval control on all three image metrics. Permutation preserves the
generated image set and therefore leaves FID unchanged, while degrading SSIM
and FeatDist. This shows that the results depend on tile-specific expression
rather than PixCell alone or retrieval of similar training examples
(\cref{sec:supp-nullcond}).

\paragraph{Which mapper components matter?}
On the same holdouts, we ablate the positional embedding, graph-attention
blocks, spatial pathway, and per-cell encoder. Removing either positional
encoding or graph attention worsens FID and FeatDist on both patients, while
replacing the full spatial pathway with mean pooling degrades FID further.
Overall, \ourmethod leads four of six comparisons and remains close to the
best baseline on the other two, while achieving the best FeatDist on both patients. Full results are in
\cref{sec:supp-nullcond}.

\paragraph{Choice of foundation model.}
\label{sec:exp-fm}
The mapper can target different FM embedding spaces by changing its output
dimension and decoder conditioning. Replacing UNI2-h with CONCH, UNI-v1, and
Virchow2 supports the mapper's portability across embedding spaces. We use UNI2-h because it is PixCell's native space (\cref{sec:supp-fm}).

\paragraph{\ourmethod hyperparameters.}
We report a post-hoc sensitivity analysis over $k \in \{4,8,16,32\}$ around
the fixed $k{=}8$ setting (\cref{sec:supp-k-ablation}). We also analyze the
readout decay $\gamma$ on the held-out $1024$px fields. Initializing $\gamma$ at $50$ gives the best CLIP FID on all three datasets and is used for the reported $1024$px results (\cref{sec:supp-decay}).

\paragraph{Additional results.}
The supplement also covers single-cell conditioning (\cref{sec:supp-indist-full}),
gene influence (\cref{sec:supp-enrichment}) and qualitative galleries
(\cref{sec:supp-galleries}).

\section{Conclusion}
\label{sec:conclusion}

Our experiments show that spatial gene expression can be mapped to
pathology foundation-model representations for efficient histology synthesis
across gene panels, image encoders, and resolutions. Separating molecular
prediction from rendering allows mapper and decoder errors to be analyzed
separately, while statistically generated expression enables paired synthetic
expression-image generation. Our generalization experiments show that the
formulation remains informative across multiple shifts, though performance
degrades under harder cohort changes. Expression-conditioned histology
synthesis remains an open problem, but these results establish
foundation-model embedding prediction as a viable path toward further
advances in molecular encoding, spatial modeling, and rendering.

\section*{Acknowledgements}
This research was supported by National Science Foundation (NSF) grants IIS-2123920, IIS-2212046.

{
    \small
    \bibliographystyle{ieeenat_fullname}
    \bibliography{main}

@inproceedings{geneflow,
  title     = {{GeneFlow}: Translation of Single-cell Gene Expression to Histopathological Images via Rectified Flow},
  author    = {Wang, Mengbo and Verma, Shourya and Malusare, Aditya and Wang, Luopin and Lu, Yiyang and Aggarwal, Vaneet and Sola, Mario and Grama, Ananth and Lanman, Nadia Atallah},
  booktitle = {Advances in Neural Information Processing Systems (NeurIPS)},
  volume    = {38},
  year      = {2025}
}

@inproceedings{ge2hist,
  title     = {{GE2Hist}: Generating Histology Images from Single-cell Gene Expression via Cross-modal Generative Network},
  author    = {Cai, Hongmin and Ji, Boan and Cai, Shangyan and Liao, Yi and Chen, Jiazhou and Huang, Weitian},
  booktitle = {Medical Image Computing and Computer Assisted Intervention (MICCAI)},
  pages     = {240--250},
  publisher = {Springer Nature Switzerland},
  year      = {2025},
  doi       = {10.1007/978-3-032-05141-7_24}
}

@article{st2he,
  title   = {{ST2HE}: enhancing spatial transcriptomics interpretability via virtual staining for histological annotation},
  author  = {Liu, Zhentao and Das, Arun and Meng, Wen and Chiu, Yu-Chiao and Gao, Shou-Jiang and Huang, Yufei},
  journal = {Briefings in Bioinformatics},
  volume  = {27},
  number  = {3},
  pages   = {bbag326},
  year    = {2026},
  doi     = {10.1093/bib/bbag326}
}

@article{uni,
  title   = {Towards a general-purpose foundation model for computational pathology},
  author  = {Chen, Richard J. and Ding, Tong and Lu, Ming Y. and Williamson, Drew F. K. and others},
  journal = {Nature Medicine},
  volume  = {30},
  number  = {3},
  pages   = {850--862},
  year    = {2024}
}

@misc{uni2,
  title        = {{UNI2-h} model card},
  author       = {{MahmoodLab}},
  howpublished = {\url{https://huggingface.co/MahmoodLab/UNI2-h}},
  year         = {2025}
}

@article{conch,
  title   = {A visual-language foundation model for computational pathology},
  author  = {Lu, Ming Y. and Chen, Bowen and Zhang, Andrew and Williamson, Drew F. K. and others},
  journal = {Nature Medicine},
  volume  = {30},
  number  = {3},
  pages   = {863--874},
  year    = {2024}
}

@article{virchow2,
  title   = {{Virchow2}: Scaling Self-Supervised Mixed Magnification Models in Pathology},
  author  = {Zimmermann, Eric and Vorontsov, Eugene and Viret, Julian and Casson, Adam and Zelechowski, Michal and Shaikovski, George and Tenenholtz, Neil and Hall, James and Klimstra, David and Yousfi, Razik and Fuchs, Thomas and Fusi, Nicolo and Liu, Siqi and Severson, Kristen},
  journal = {arXiv preprint arXiv:2408.00738},
  year    = {2024}
}

@article{pixcell,
  title   = {{PixCell}: A generative foundation model for digital histopathology images},
  author  = {Yellapragada, Srikar and Graikos, Alexandros and Li, Zilinghan and Triaridis, Kostas and Belagali, Varun and Nandi, Tarak Nath and Bai, Karen and Knudsen, Beatrice S. and Kurc, Tahsin and Gupta, Rajarsi R. and Prasanna, Prateek and Madduri, Ravi K. and Saltz, Joel and Samaras, Dimitris},
  journal = {arXiv preprint arXiv:2506.05127},
  year    = {2025}
}

@inproceedings{graikos2024learned,
  title={Learned representation-guided diffusion models for large-image generation},
  author={Graikos, Alexandros and Yellapragada, Srikar and Le, Minh-Quan and Kapse, Saarthak and Prasanna, Prateek and Saltz, Joel and Samaras, Dimitris},
  booktitle={2024 IEEE/CVF Conference on Computer Vision and Pattern Recognition (CVPR)},
  pages={8532--8542},
  year={2024},
  organization={IEEE}
}

@article{spirit,
  title   = {Inferring single-cell spatial gene expression with tissue morphology via explainable deep learning},
  author  = {Zhao, Yue and Alizadeh, Elaheh and Taha, Hash Brown and Liu, Yang and Xu, Ming and Mahoney, J. Matthew and Li, Sheng},
  journal = {bioRxiv preprint 2024.06.12.598686},
  year    = {2025}
}

@article{ghist,
  title   = {Spatial gene expression at single-cell resolution from histology using deep learning with {GHIST}},
  author  = {Fu, Xiaohang and Cao, Yue and Bian, Beilei and Wang, Chuhan and Graham, Dinny and Pathmanathan, Nirmala and Patrick, Ellis and Kim, Jinman and Yang, Jean Yee Hwa},
  journal = {Nature Methods},
  volume  = {22},
  pages   = {1900--1910},
  year    = {2025},
  doi     = {10.1038/s41592-025-02795-z}
}

@inproceedings{stem,
  title     = {Diffusion Generative Modeling for Spatially Resolved Gene Expression Inference from Histology Images},
  author    = {Zhu, Sichen and Zhu, Yuchen and Tao, Molei and Qiu, Peng},
  booktitle = {International Conference on Learning Representations (ICLR)},
  year      = {2025}
}

@article{omiclip,
  title   = {A visual--omics foundation model to bridge histopathology with spatial transcriptomics},
  author  = {Chen, Weiqing and Zhang, Pengzhi and Tran, Tu N. and Xiao, Yiwei and Li, Shengyu and Shah, Vrutant V. and Cheng, Hao and Brannan, Kristopher W. and Youker, Keith and Lai, Li and Fang, Longhou and Yang, Yu and Le, Nhat-Tu and Abe, Jun-ichi and Chen, Shu-Hsia and Ma, Qin and Chen, Ken and Song, Qianqian and Cooke, John P. and Wang, Guangyu},
  journal = {Nature Methods},
  volume  = {22},
  pages   = {1568--1582},
  year    = {2025},
  doi     = {10.1038/s41592-025-02707-1}
}

@article{storm,
  title   = {A Multimodal Foundation Model of Spatial Transcriptomics and Histology for Biological Discovery and Clinical Prediction},
  author  = {Xiang, Jinxi and Hou, Siyu and Li, Yuchen and Quinton, Ryan and Zhang, Xiaoming and Eweje, Feyisope and Luo, Xiangde and Chen, Yijiang and Li, Zhe and Bergstrom, Colin and Kim, Ted and Willens, Sierra and Olguin, Francesca Maria and Abikenari, Matthew and Heider, Andrew and Rajaram, Sanjeeth and Neal, Joel and Diehn, Maximilian and Zhou, Xiang and Li, Ruijiang},
  journal = {arXiv preprint arXiv:2604.03630},
  year    = {2026}
}

@article{henium,
  title   = {{H\&Enium}, Applying Foundation Models to Computational Pathology and Spatial Transcriptomics to Learn an Aligned Latent Space},
  author  = {Glettig, Marc and Ehrensperger, Tim and Yates, Josephine and Boeva, Valentina},
  journal = {bioRxiv preprint 2025.07.22.665986},
  year    = {2025}
}

@article{cellvit,
  title   = {{CellViT}: Vision Transformers for precise cell segmentation and classification},
  author  = {H{\"o}rst, Fabian and Rempe, Moritz and Heine, Lukas and Seibold, Constantin
             and Keyl, Julius and Baldini, Giulia and Ugurel, Selma and Siveke, Jens and
             Gr{\"u}nwald, Barbara and Egger, Jan and Kleesiek, Jens},
  journal = {Medical Image Analysis},
  volume  = {94},
  pages   = {103143},
  year    = {2024},
  doi     = {10.1016/j.media.2024.103143}
}

@article{duboudin2026cytosyn,
  title={{CytoSyn}: a Foundation Diffusion Model for Histopathology--Tech Report},
  author={Duboudin, Thomas and Fontaine, Xavier and Andrier, Etienne and Guillou, Lionel and Filiot, Alexandre and Baiocco-Rodrigues, Thalyssa and Olivier, Antoine and Romagnoni, Alberto and Klein, John and Schiratti, Jean-Baptiste},
  journal={arXiv preprint arXiv:2603.18089},
  year={2026}
}

@article{xiang2026generative,
  title={A Generative Foundation Model for Multimodal Histopathology},
  author={Xiang, Jinxi and Li, Mingjie and Hou, Siyu and Chen, Yijiang and Luo, Xiangde and Ji, Yuanfeng and Zhou, Xiang and Adeli, Ehsan and Chaudhari, Akshay and Langlotz, Curtis P. and Pohl, Kilian M. and Li, Ruijiang},
  journal={arXiv preprint arXiv:2604.03635},
  year={2026}
}

@article{chen2023spatial,
  title={Spatial transcriptomic technologies},
  author={Chen, Tsai-Ying and You, Li and Hardillo, Jose Angelito U and Chien, Miao-Ping},
  journal={Cells},
  volume={12},
  number={16},
  pages={2042},
  year={2023},
  publisher={MDPI},
  doi={10.3390/cells12162042}
}

@article{tian2023expanding,
  title={The expanding vistas of spatial transcriptomics},
  author={Tian, Luyi and Chen, Fei and Macosko, Evan Z},
  journal={Nature biotechnology},
  volume={41},
  number={6},
  pages={773--782},
  year={2023},
  publisher={Nature Publishing Group US New York},
  doi={10.1038/s41587-022-01448-2}
}

@article{i2gbench,
  title   = {Performance of deep learning methods for spatial gene expression prediction from histology images: a comprehensive assessment},
  author  = {Xie, Jacky and Tan, Xiao and Jiang, Yuanhao and Nguyen, Hao and Ye, Nan and Nguyen, Quan},
  journal = {Bioinformatics Advances},
  volume  = {6},
  number  = {1},
  pages   = {vbag202},
  year    = {2026},
  doi     = {10.1093/bioadv/vbag202}
}

@article{scdesign3,
  title   = {{scDesign3} generates realistic in silico data for multimodal single-cell and spatial omics},
  author  = {Song, Dongyuan and Wang, Qingyang and Yan, Guanao and Liu, Tianyang and Sun, Tianyi and Li, Jingyi Jessica},
  journal = {Nature Biotechnology},
  volume  = {42},
  pages   = {247--252},
  year    = {2024},
  doi     = {10.1038/s41587-023-01772-1}
}

@inproceedings{hu2022lora,
  title     = {{LoRA}: Low-Rank Adaptation of Large Language Models},
  author    = {Hu, Edward J. and Shen, Yelong and Wallis, Phillip and Allen-Zhu, Zeyuan and Li, Yuanzhi and Wang, Shean and Wang, Lu and Chen, Weizhu},
  booktitle = {International Conference on Learning Representations (ICLR)},
  year      = {2022},
  url       = {https://openreview.net/forum?id=nZeVKeeFYf9}
}

@article{sccube,
  title   = {Simulating multiple variability in spatially resolved transcriptomics with {scCube}},
  author  = {Qian, Jingyang and Bao, Hudong and Shao, Xin and Fang, Yin and Liao, Jie and Chen, Zhuo and Li, Chengyu and Guo, Wenbo and Hu, Yining and Li, Anyao and Yao, Yue and Fan, Xiaohui and Cheng, Yiyu},
  journal = {Nature Communications},
  volume  = {15},
  pages   = {5021},
  year    = {2024},
  doi     = {10.1038/s41467-024-49445-0}
}

@article{sccigen,
  title   = {{sCCIgen}: a high-fidelity spatially resolved transcriptomics data simulator for cell--cell interaction studies},
  author  = {Song, Xiaoyu and Chavez-Fuentes, Joselyn C. and Ma, Weiping and Fu, Weijia and Shin, Sujung Crystal and Wang, Pei and Yuan, Guo-Cheng},
  journal = {Genome Biology},
  volume  = {26},
  pages   = {354},
  year    = {2025},
  doi     = {10.1186/s13059-025-03762-9}
}

@book{cressie1993spatial,
  title     = {Statistics for Spatial Data},
  author    = {Cressie, Noel A. C.},
  publisher = {Wiley},
  edition   = {Revised},
  year      = {1993}
}

@book{diggle2013point,
  title     = {Statistical Analysis of Spatial and Spatio-Temporal Point Patterns},
  author    = {Diggle, Peter J.},
  publisher = {Chapman and Hall/CRC},
  edition   = {Third},
  year      = {2013}
}

@inproceedings{hest,
  title     = {{HEST-1k}: A Dataset for Spatial Transcriptomics and Histology Image Analysis},
  author    = {Jaume, Guillaume and Doucet, Paul and Song, Andrew H. and Lu, Ming Y. and Almagro-P{\'e}rez, Cristina and Wagner, Sophia J. and Vaidya, Anurag J. and Chen, Richard J. and Williamson, Drew F. K. and Kim, Ahrong and Mahmood, Faisal},
  booktitle = {Advances in Neural Information Processing Systems},
  volume    = {37},
  year      = {2024},
  doi       = {10.52202/079017-1704},
  url       = {https://proceedings.neurips.cc/paper_files/paper/2024/hash/60a899cc31f763be0bde781a75e04458-Abstract-Datasets_and_Benchmarks_Track.html}
}

@inproceedings{cleanfid,
  title     = {On Aliased Resizing and Surprising Subtleties in {GAN} Evaluation},
  author    = {Parmar, Gaurav and Zhang, Richard and Zhu, Jun-Yan},
  booktitle = {Proceedings of the IEEE/CVF Conference on Computer Vision and Pattern Recognition},
  pages     = {11410--11420},
  year      = {2022}
}

@inproceedings{kong2025spatia,
  title={{SPATIA}: Multimodal Generation and Prediction of Spatial Cell Phenotypes},
  author={Kong, Zhenglun and Qiu, Mufan and Boesen, John and Lin, Xiang and Yun, Sukwon and Chen, Tianlong and Kellis, Manolis and Zitnik, Marinka},
  booktitle={International Conference on Machine Learning (ICML)},
  year={2026}
}

@inproceedings{vlachas2026transcriptomics,
  title={Transcriptomics-Conditioned Virtual Tissue Synthesis via Diffusion Transformers},
  author={Vlachas, Pantelis R and Nonchev, Kalin and Koelzer, Viktor H and R{\"a}tsch, Gunnar},
  booktitle={ICML Workshop on Structured Data for Health (SD4H)},
  year={2026}
}

@inproceedings{
liu2023flow,
title={Flow Straight and Fast: Learning to Generate and Transfer Data with Rectified Flow},
author={Xingchao Liu and Chengyue Gong and Qiang Liu},
booktitle={The Eleventh International Conference on Learning Representations },
year={2023},
url={https://openreview.net/forum?id=XVjTT1nw5z}
}

@inproceedings{yellapragada2025zoomldm,
  title={{ZoomLDM}: Latent diffusion model for multi-scale image generation},
  author={Yellapragada, Srikar and Graikos, Alexandros and Triaridis, Kostas and Prasanna, Prateek and Gupta, Rajarsi and Saltz, Joel and Samaras, Dimitris},
  booktitle={2025 IEEE/CVF Conference on Computer Vision and Pattern Recognition (CVPR)},
  pages={23453--23463},
  year={2025},
  organization={IEEE}
}

@inproceedings{le2026construct,
  title     = {{ConStruct}: Structural Distillation of Foundation Models
               for Prototype-Based Weakly Supervised Histopathology Segmentation},
  author    = {Le, Khang and Thach, Ha and Vu, Anh M. and Vo, Trang T. K.
               and Huynh, Han H. and Yang, David and Le, Minh H. N.
               and Nguyen, Thanh-Huy and Awasthi, Akash and Mohan, Chandra
               and Han, Zhu and Nguyen, Hien Van},
  booktitle = {Proceedings of the 9th International Conference on
               Medical Imaging with Deep Learning (MIDL)},
  series    = {Proceedings of Machine Learning Research},
  volume    = {315},
  pages     = {2893--2905},
  year      = {2026},
  publisher = {PMLR}
}

@inproceedings{pham2026disentangled,
  title     = {Learning Disentangled Stain and Structural Representations
               for Semi-Supervised Histopathology Segmentation},
  author    = {Pham, Ha-Hieu and Vu, Nguyen Lan Vi and Nguyen, Thanh-Huy
               and Bagci, Ulas and Xu, Min and Le, Trung-Nghia
               and Pham, Huy-Hieu},
  booktitle = {Proceedings of the MICCAI Workshop on Computational Pathology},
  series    = {Proceedings of Machine Learning Research},
  volume    = {316},
  pages     = {278--287},
  year      = {2026},
  publisher = {PMLR}
}
}

\clearpage
\appendix
\begin{center}{\Large\bf
    GESTALT: Fast, Generalizable Histopathology Image Synthesis from
    Gene Expression with Pretrained Pathology Foundation Models \\ \phantom{a} \\
    Supplementary Material}
\end{center}
\suppressfloats[t]
\newcommand{\mainpaperref}[2]{\cref{#1}}
\setcounter{section}{0}
\setcounter{table}{0}
\setcounter{figure}{0}
\setcounter{equation}{0}
\renewcommand{\thesection}{S\arabic{section}}
\renewcommand{\thetable}{S\arabic{table}}
\renewcommand{\thefigure}{S\arabic{figure}}
\renewcommand{\theequation}{S\arabic{equation}}

\renewcommand{\theHsection}{S\arabic{section}}
\renewcommand{\theHtable}{S\arabic{table}}
\renewcommand{\theHfigure}{S\arabic{figure}}
\renewcommand{\theHequation}{S\arabic{equation}}

\section{Separating mapper and renderer error}
\label{sec:supp-lora}
The adapter never uses expression or mapper predictions: it is fitted only on
real $(e,I)$ pairs from the training split. Held-out tiles are excluded; in
zero-shot experiments, all evaluation slides are excluded from both
mapper and adapter training.

For a held-out tile $I$, the same decoder renders our prediction $D(\hat e)$
and the oracle $D(e)$. Oracle metrics show the error left with the real
embedding; the prediction-to-oracle gap shows the additional effect of using
$\hat e$. $\unicos=\operatorname{cos}(\hat e,e)$ measures this difference
before decoding.

As the test setting gets harder, the main source of error moves from the
decoder to the mapper. From
in-distribution C1/C2/P1 to cross-condition lung and cross-study breast, the FID gap increases
from $1.5$--$3.3$ to $15.8$--$42.6$ and $62.5$--$76.8$, while $\unicos$ falls
from $0.81$--$0.86$ to $0.66$--$0.73$ and $0.40$--$0.48$. Thus, much of the
remaining in-distribution error is already present at the decoder, while
embedding transfer becomes limiting under distribution shift. The trend is
descriptive because the protocols also differ in dataset, resolution and
evaluation size.

\section{Xenium C1, C2 and P1}
\label{sec:supp-indist-full}
\subsection{\texorpdfstring{Repeated-run results at native $40\times$}{Repeated-run results at native 40x}}
The main paper reports means over three independently trained mappers and
LoRAs. \Cref{tab:supp-indist-seeds} gives the corresponding standard
deviations on the same fixed held-out tiles alongside \geneflow's published
results.

\begin{figure*}[t]
\centering
\includegraphics[width=\linewidth]{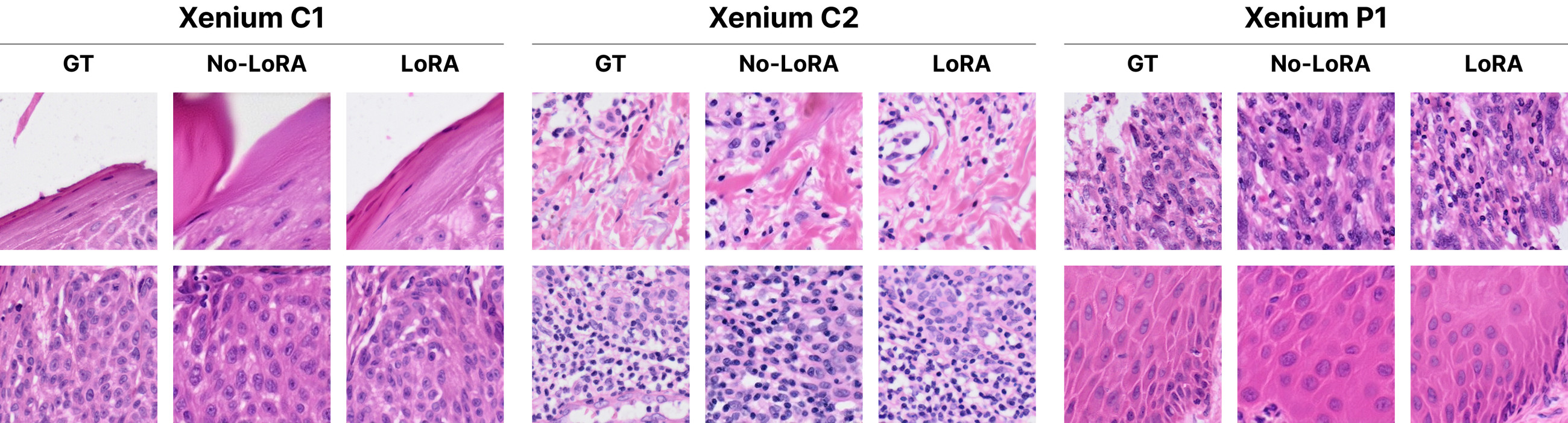}
\caption{\textbf{Effect of decoder adaptation at native $20\times$.}
Each dataset shows two held-out tiles spanning lower to higher cell density.
Columns compare the ground truth, the mapper prediction rendered by frozen
PixCell (No-LoRA), and the same prediction rendered by the adapted decoder
(LoRA). Adaptation improves agreement in tissue extent and nuclear morphology.}
\label{fig:supp-qualitative-20x}
\end{figure*}

The reported gain is stable across independent runs: FID standard deviation is
only $0.08$--$0.17$, against a $21$--$32$ point improvement over \geneflow.
The comparison is therefore not driven by one run.

\subsection{Single-cell conditioning}
\label{sec:supp-singlecell}
\geneflow~\cite{geneflow} evaluates two settings. Multi-cell conditioning
uses every cell in a fixed tissue window. Single-cell conditioning centers a
window on one cell and uses only that cell's expression.
The main paper compares against \geneflow's multi-cell numbers
(\mainpaperref{tab:indist}{Table~1 of the main paper}).
\Cref{tab:supp-singlecell} reports the single-cell setting, both sides at
native $40\times$.
The advantage is not specific to multi-cell conditioning: in the single-cell
setting, \ourmethod lowers mean FID by $2.6$--$24.9$ and mean FeatDist by
$0.9$--$3.1$ on all three, while improving mean SSIM on 2/3.

\begin{table}[h]
\centering\small
\setlength{\tabcolsep}{10pt}
\resizebox{\linewidth}{!}{
\begin{tabular}{lccc}
\toprule
 & FID $\downarrow$ & SSIM $\uparrow$ & FeatDist $\downarrow$ \\
\midrule
\multicolumn{4}{l}{\emph{\geneflow~\cite{geneflow} (single-cell, reported)}} \\
Xenium$_{C1}$ & $20.73{\pm}8.45$ & $0.24{\pm}0.04$ & $14.90{\pm}2.55$ \\
Xenium$_{C2}$ & $42.61{\pm}4.50$ & $\mathbf{0.35{\pm}0.06}$ & $15.65{\pm}2.28$ \\
Xenium$_{P1}$ & $27.43{\pm}5.91$ & $0.17{\pm}0.03$ & $14.53{\pm}2.15$ \\
\midrule
\multicolumn{4}{l}{\emph{\ourmethod (ours, LoRA, single-cell)}} \\
Xenium$_{C1}$ & $\mathbf{18.16{\pm}2.54}$ & $\mathbf{0.3202{\pm}0.0003}$ & $\mathbf{13.38{\pm}0.09}$ \\
Xenium$_{C2}$ & $\mathbf{17.74{\pm}2.41}$ & $0.3263{\pm}0.0019$ & $\mathbf{12.52{\pm}0.05}$ \\
Xenium$_{P1}$ & $\mathbf{20.01{\pm}1.20}$ & $\mathbf{0.2284{\pm}0.0010}$ & $\mathbf{13.63{\pm}0.08}$ \\
\bottomrule
\end{tabular}}
\caption{Single-cell conditioning, both sides at native $40\times$. GESTALT
reports mean $\pm$ s.d. over three independently trained mapper and decoder
LoRA runs on the same fixed full holdout; \geneflow values are as
reported. GESTALT improves FID and FeatDist on all three datasets and SSIM on
two of three, the exception being SSIM on Xenium$_{C2}$. Every held-out cell is
evaluated in each GESTALT run, $20\,113$ / $15\,748$ / $19\,584$ for C1 / C2 /
P1.}
\label{tab:supp-singlecell}
\end{table}

\subsection{\texorpdfstring{Native $20\times$ results}{Native 20x results}}
\Cref{tab:supp-indist-full} reports our in-distribution results at PixCell's
native $20\times$ resolution (0.5\,$\mu$m/px). The main paper uses
$40\times$ to match \geneflow's evaluation.
The pipeline therefore also works at PixCell's native resolution. These results are not
directly comparable with the $40\times$ since the mapper and adapter
are trained separately at each resolution.

\Cref{fig:supp-qualitative-20x} shows that frozen PixCell produces plausible
H\&E appearance but often mismatches tissue extent and nuclear morphology;
LoRA better follows the ground truth.

\begin{table}[ht]
\centering\small
\begin{tabular}{lccc}
\toprule
 & FID $\downarrow$ & SSIM $\uparrow$ & FeatDist $\downarrow$ \\
\midrule
Xenium$_{C1}$ & 28.7 & 0.22 & 12.6 \\
Xenium$_{C2}$ & 29.8 & 0.26 & 12.0 \\
Xenium$_{P1}$ & 24.3 & 0.19 & 12.1 \\
\bottomrule
\end{tabular}
\caption{In-distribution results for the LoRA-adapted pipeline at PixCell's
native $20\times$. The mapper and adapter are trained at this resolution, so
these values are not directly compared with the separately trained
$40\times$ runs.}
\label{tab:supp-indist-full}
\end{table}

\begin{table*}[ht]
\centering\small
\begin{tabular}{llcccc}
\toprule
Dataset & Conditioning & FID $\downarrow$ & SSIM $\uparrow$ & FeatDist $\downarrow$
& \unicos $\uparrow$ \\
\midrule
Xenium$_{C1}$ & \geneflow
& $47.95{\pm}7.38$ & $0.28{\pm}0.10$ & $14.50{\pm}2.80$ & -- \\
& Ours
& $\mathbf{15.75{\pm}0.08}$ & $\mathbf{0.3500{\pm}0.0004}$ & $\mathbf{12.66{\pm}0.01}$ & $0.838{\pm}0.001$ \\
& Oracle
& \emph{14.27$\pm$0.06} & \emph{0.3525$\pm$0.0005} & \emph{11.94$\pm$0.01} & -- \\
\midrule
Xenium$_{C2}$ & \geneflow
& $45.50{\pm}4.10$ & $\mathbf{0.40{\pm}0.09}$ & $15.61{\pm}2.26$ & -- \\
& Ours
& $\mathbf{18.03{\pm}0.11}$ & $0.3964{\pm}0.0005$ & $\mathbf{12.47{\pm}0.07}$ & $0.860{\pm}0.001$ \\
& Oracle
& \emph{14.75$\pm$0.04} & \emph{0.3996$\pm$0.0006} & \emph{11.69$\pm$0.05} & -- \\
\midrule
Xenium$_{P1}$ & \geneflow
& $34.31{\pm}5.65$ & $0.23{\pm}0.11$ & $13.41{\pm}2.19$ & -- \\
& Ours
& $\mathbf{13.15{\pm}0.17}$ & $\mathbf{0.2829{\pm}0.0007}$ & $\mathbf{12.52{\pm}0.02}$ & $0.807{\pm}0.001$ \\
& Oracle
& \emph{11.54$\pm$0.18} & \emph{0.2870$\pm$0.0004} & \emph{11.73$\pm$0.03} & -- \\
\bottomrule
\end{tabular}
\caption{In-distribution results at native $40\times$. GESTALT and oracle rows
report mean $\pm$ s.d. over three independent runs on the same fixed held-out
split; \geneflow~\cite{geneflow} values are reproduced as reported. Each
GESTALT run independently initializes and trains the mapper and decoder LoRA
and sets the diffusion sampling noise.}
\label{tab:supp-indist-seeds}
\end{table*}

\section{The $1024{\times}1024$ grid readout}
\label{sec:supp-grid1024}

We complement the main-paper definition of the grid readout by testing whether
its sixteen queries remain local while the shared backbone communicates across
sub-region boundaries.

\begin{figure}[t]
\centering
\includegraphics[width=\linewidth]{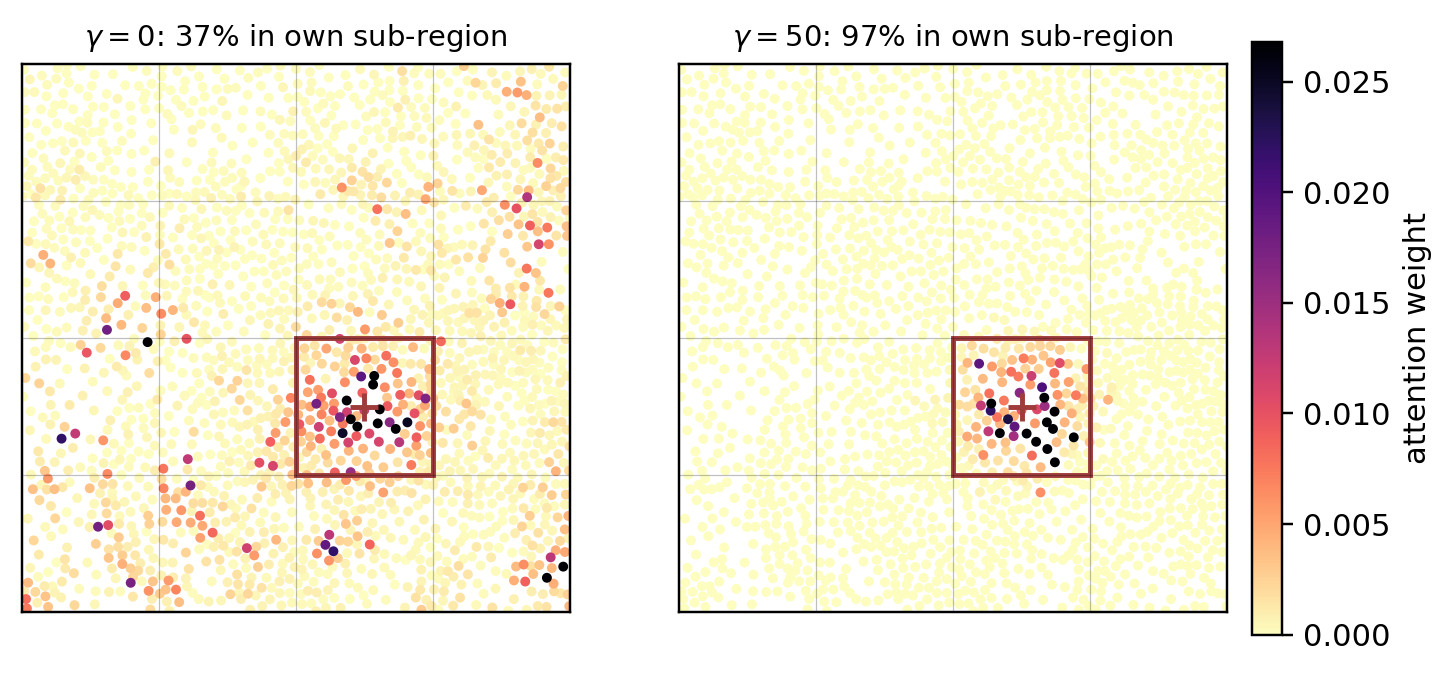}
\caption{\textbf{Readout attention of a single query} over a held-out C1
field, with its sub-region outlined. The $\gamma{=}50$ penalty raises this
query's in-region attention from $37\%$ (left) to $97\%$ (right). These are
values for one query, not dataset means; the per-dataset means over all
sixteen queries are given in the text of this section.}
\label{fig:supp-grid1024-attn}
\end{figure}
The distance penalty makes the readout local (\cref{fig:supp-grid1024-attn}):
at $\gamma{=}50$, attention within a query's own sub-region rises from
$41.4\%$ to $90.3\%$ on C1, $50.9\%$ to $91.5\%$ on C2 and $38.1\%$ to
$88.4\%$ on P1. Increasing $\gamma$ to $100$ reaches only $88$--$95\%$, so
locality is already near saturation at $50$. The penalty affects only the
readout; the shared backbone continues to pass context across boundaries.

\paragraph{Choosing the readout decay.}
\label{sec:supp-decay}
\Cref{tab:supp-decay} sweeps the initial value of the learnable $\gamma$, the
hyperparameter introduced by the grid readout; $\gamma{=}0$ disables the
penalty.
Either positive penalty improves both metrics over $\gamma{=}0$ on all three
datasets (\cref{tab:supp-decay}). We use $\gamma{=}50$ because it gives the
best CLIP FID among the swept settings on all three and the best Crop FID on
C2 and P1; $\gamma{=}100$ leads only Crop FID on C1.

\begin{figure*}[!t]
\centering
\includegraphics[width=0.95\linewidth]{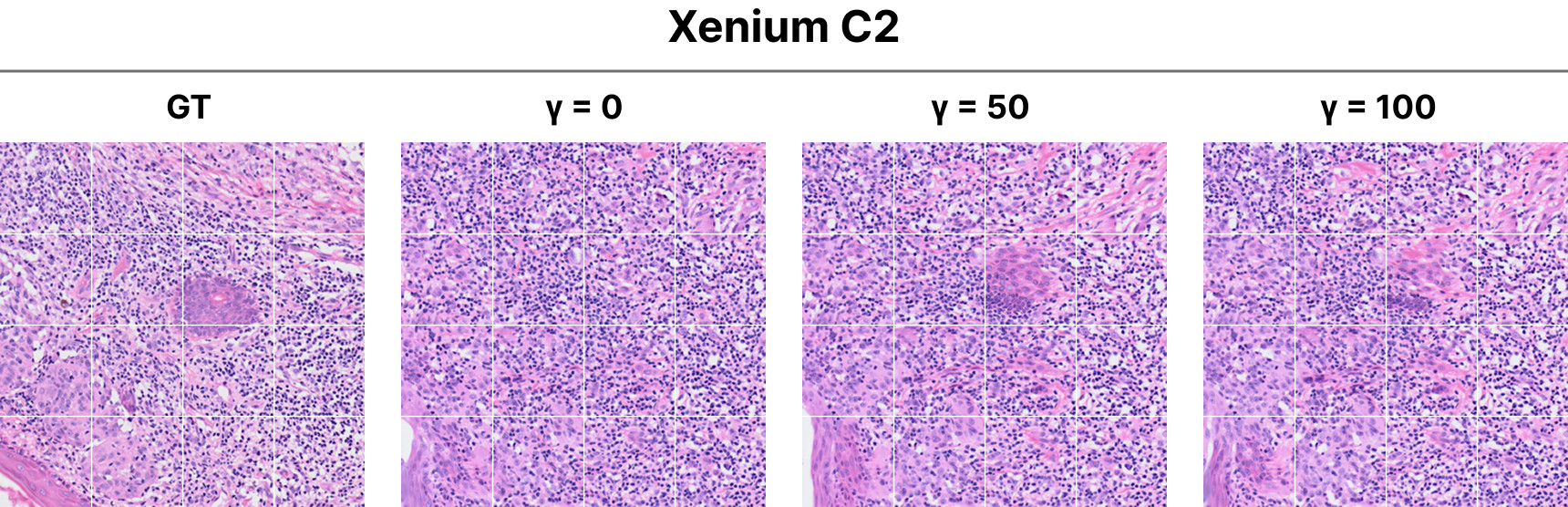}
\caption{\textbf{Readout decay on held-out $1024{\times}1024$ fields.}
Columns compare the ground truth with $\gamma\in\{0,50,100\}$; grid lines mark
the sixteen readout regions. With $\gamma{=}0$, regional variation collapses
and pale or empty areas fill with tissue. Both positive settings recover these
structures. Quantitative Crop FID / CLIP FID results are reported in
\cref{tab:supp-decay}.}
\label{fig:supp-decay-gallery}
\end{figure*}

The failure mode is visible in \cref{fig:supp-decay-gallery}: without the
penalty, the sixteen outputs become too similar and overwrite pale or empty
regions; either positive setting restores regional variation.

\begin{table}[h]
\centering\small
\setlength{\tabcolsep}{5pt}
\begin{tabular}{llcc}
\toprule
& & Crop FID $\downarrow$ & CLIP FID $\downarrow$ \\
\midrule
\multirow{4}{*}{Xenium$_{C1}$}
 & $\gamma{=}0$   & 75.96 & 8.54 \\
 & $\gamma{=}50$  & 71.59 & \textbf{5.29} \\
 & $\gamma{=}100$ & \textbf{66.37} & 5.84 \\
 & \emph{Oracle}  & \emph{59.12} & \emph{3.62} \\
\midrule
\multirow{4}{*}{Xenium$_{C2}$}
 & $\gamma{=}0$   & 63.93 & 5.70 \\
 & $\gamma{=}50$  & \textbf{50.87} & \textbf{3.33} \\
 & $\gamma{=}100$ & 60.81 & 3.51 \\
 & \emph{Oracle}  & \emph{39.20} & \emph{2.08} \\
\midrule
\multirow{4}{*}{Xenium$_{P1}$}
 & $\gamma{=}0$   & 56.75 & 4.13 \\
 & $\gamma{=}50$  & \textbf{49.87} & \textbf{2.98} \\
 & $\gamma{=}100$ & 51.40 & 3.37 \\
 & \emph{Oracle}  & \emph{47.92} & \emph{2.91} \\
\bottomrule
\end{tabular}
\caption{\textbf{Readout decay sweep} on all fully covered held-out fields.
Settings share the adapter, seed and sampler noise, so they differ only in
the readout penalty.}
\label{tab:supp-decay}
\end{table}

\section{Conditioning and mapper design}
\label{sec:supp-nullcond}
\providecommand{\ablscale}{0.8}
\paragraph{Why ablate on HEST lung?}
C1/C2/P1 provide the matched comparison with \geneflow, but their
in-distribution protocol holds out tiles rather than slides. Training and test
tiles may therefore share slide-specific appearance, making that setting less
diagnostic of whether the conditioning and mapper generalize. HEST lung
instead withholds both evaluation patients and the healthy target state from
training. We use this stricter setting to test whether the result depends on
tile-specific expression and whether the mapper needs spatial and
cell-resolved components. Every comparison uses the same training pool,
decoder adapter, and full NCBI884 and NCBI875 evaluation slides. Full
\ourmethod and oracle results on these slides in
\cref{tab:supp-lung-crosscond}.

\begin{table}[h]
\centering\small
\setlength{\tabcolsep}{4pt}
\begin{tabular}{lcccc}
\toprule
 & FID $\downarrow$ & SSIM $\uparrow$ & FeatDist $\downarrow$ & \unicos $\uparrow$ \\
\midrule
\multicolumn{5}{l}{\emph{NCBI884 (unseen healthy patient)}} \\
\ourmethod & 45.86 & 0.203 & 13.70 & 0.732 \\
\emph{Oracle} & 30.07 & 0.221 & 11.98 & -- \\
\midrule
\multicolumn{5}{l}{\emph{NCBI875 (unseen healthy patient)}} \\
\ourmethod & 89.65 & 0.313 & 14.31 & 0.659 \\
\emph{Oracle} & 47.08 & 0.324 & 11.91 & -- \\
\bottomrule
\end{tabular}
\caption{\textbf{Cross-condition HEST lung holdouts.} Training is on
15 diseased slides, so each holdout crosses patient and disease state
together. Full slides, $n=708$ and $n=379$, one checkpoint and one adapter.
The mapper sits $15.8$ FID above the oracle on NCBI884, $42.6$ on NCBI875.}
\label{tab:supp-lung-crosscond}
\end{table}

\paragraph{Does expression conditioning matter?} Keeping the decoder fixed, we
compare three alternatives. A constant training-set mean tests PixCell without
tile-specific information. Permutation gives each tile another tile's
predicted embedding and diffusion seed, so the set of generated images is
unchanged but each image no longer belongs to its own tile. Top-10 retrieval tests whether similar training expression suffices without a
learned mapper.

\begin{table}[h]
\centering\small
\resizebox{\ablscale\columnwidth}{!}{%
\begin{tabular}{lccc}
\toprule
 & FID $\downarrow$ & SSIM $\uparrow$ & FeatDist $\downarrow$ \\
\midrule
\multicolumn{4}{l}{\emph{NCBI884 (unseen healthy patient)}} \\
Constant mean       & 86.47 & 0.142 & 15.06 \\
Permuted prediction & 45.86 & 0.165 & 15.24 \\
Top-10 retrieval    & 69.53 & 0.139 & 14.59 \\
\ourmethod           & \textbf{45.86} & \textbf{0.203} & \textbf{13.70} \\
\emph{Oracle (reference)} & 30.07 & 0.221 & 11.98 \\
\midrule
\multicolumn{4}{l}{\emph{NCBI875 (unseen healthy patient)}} \\
Constant mean       & 113.48 & 0.169 & 15.43 \\
Permuted prediction & 89.65  & 0.257 & 15.87 \\
Top-10 retrieval    & 97.41  & 0.191 & 15.25 \\
\ourmethod           & \textbf{89.65} & \textbf{0.313} & \textbf{14.31} \\
\emph{Oracle (reference)} & 47.08 & 0.324 & 11.91 \\
\bottomrule
\end{tabular}}
\caption{\textbf{Conditioning-input ablation, HEST lung.}
All rows use the complete evaluation slides and the same decoder LoRA.
Permutation leaves the set of generated images unchanged, so its FID equals
\ourmethod's by construction.}
\label{tab:supp-nullcond}
\end{table}

\Cref{tab:supp-nullcond} supports our conditioning design. The full model beats
the constant and permutation controls on both paired metrics, and retrieval on
all three metrics, on both slides. Permutation keeps the same images, so its
FID is identical, yet both paired metrics worsen because each image no longer
matches its tile. The result therefore depends on giving each tissue field its
own molecular state, rather than on PixCell producing plausible histology
alone.

\paragraph{Which mapper components matter?} We remove one component at a time,
retraining each variant under the same protocol. No positional embedding
removes only the explicit absolute-coordinate embedding $p_i$; the $k$-NN
graph and relative-offset bias remain. No graph attention removes message
passing; Mean-pool MLP replaces the spatial pathway with mean and
standard-deviation pooling; and Detected-mean MLP also replaces the per-cell
encoder with detected-gene means and log cell count.

\begin{table}[h]
\centering\small
\resizebox{\linewidth}{!}{%
\begin{tabular}{lccc}
\toprule
 & FID $\downarrow$ & SSIM $\uparrow$ & FeatDist $\downarrow$ \\
\midrule
\multicolumn{4}{l}{\emph{NCBI884 (unseen healthy patient)}} \\
No positional embedding & 48.50 & 0.182 & 14.13 \\
No graph attention      & 46.57 & 0.218 & 14.02 \\
Mean-pool MLP           & 68.51 & \textbf{0.224} & 14.56 \\
Detected-mean MLP       & 69.28 & 0.160 & 14.34 \\
\ourmethod               & \textbf{45.86} & 0.203 & \textbf{13.70} \\
\emph{Oracle (reference)} & 30.07 & 0.221 & 11.98 \\
\midrule
\multicolumn{4}{l}{\emph{NCBI875 (unseen healthy patient)}} \\
No positional embedding & 104.22 & 0.266 & 14.96 \\
No graph attention      & 100.50 & 0.306 & 14.75 \\
Mean-pool MLP           & 127.58 & 0.308 & 15.25 \\
Detected-mean MLP       & \textbf{87.32} & 0.238 & 14.54 \\
\ourmethod               & 89.65 & \textbf{0.313} & \textbf{14.31} \\
\emph{Oracle (reference)} & 47.08 & 0.324 & 11.91 \\
\bottomrule
\end{tabular}}
\caption{\textbf{Mapper-architecture ablation on held-out HEST lung
patients.} Each architecture uses the same training pool, decoder LoRA, and
evaluation protocol on the full evaluation slides.}
\label{tab:supp-mapper-architecture}
\end{table}

Because every architecture in \cref{tab:supp-mapper-architecture} uses the
same adapted PixCell decoder, differences between rows isolate the
gene-to-embedding mapping while holding the renderer fixed. Replacing the
spatial pathway with mean pooling increases FID by $22.65$ on NCBI884 and
$37.93$ on NCBI875, showing that the full-system performance cannot be
attributed to the pretrained renderer alone. This spatial modeling also
differs from \geneflow, whose
multi-cell encoder aggregates per-cell expression without explicitly using
cell coordinates~\cite{geneflow}; \ourmethod instead incorporates coordinates
through its positional and neighborhood-based spatial pathway.
Removing either the absolute-position embedding or
graph attention worsens FID and FeatDist on both slides, with the larger loss
from removing the position embedding, while removing the per-cell encoder
lowers SSIM on both.
\ourmethod leads four of six metrics: FID and FeatDist on NCBI884, and SSIM
and FeatDist on NCBI875. Mean-pool MLP leads NCBI884 SSIM but incurs much
higher FID and FeatDist, while Detected-mean MLP leads NCBI875 FID but sharply
reduces SSIM. Together, these results favor the spatial, cell-resolved
mapper, with no single metric driving the conclusion.

\section{\texorpdfstring{Sensitivity to $k$-NN neighborhood size}{Sensitivity to k-NN neighborhood size}}
\label{sec:supp-k-ablation}
The main paper fixes $k{=}8$. \Cref{tab:supp-kablation} varies only the
mapper's neighborhood size on the two cross-condition lung holdouts; all settings share the
same LoRA-adapted decoder.
By FID, $k{=}8$ is the best cross-slide compromise: it ranks first on NCBI875,
$0.65$ ahead of $k{=}32$, and second on NCBI884, only $0.21$ behind $k{=}4$.
SSIM spans at most $0.010$ on either slide. This post-hoc result supports
$k{=}8$ as the fixed default.

\begin{table}[h]
\centering\small
\begin{tabular}{lcccc}
\toprule
 & \multicolumn{2}{c}{NCBI875} & \multicolumn{2}{c}{NCBI884} \\
$k$ & FID $\downarrow$ & SSIM $\uparrow$ & FID $\downarrow$ & SSIM $\uparrow$ \\
\midrule
4  & 107.75 & \textbf{0.316} & \textbf{45.65} & \textbf{0.210} \\
8  & \textbf{89.65} & 0.313 & 45.86 & 0.203 \\
16 & 97.36 & \textbf{0.316} & 47.73 & 0.206 \\
32 & 90.30  & 0.306 & 48.40 & 0.202 \\
\bottomrule
\end{tabular}
\caption{Post-hoc sensitivity to $k$ on the two cross-condition lung holdouts. Both slides are
excluded from mapper and adapter training, and all settings share the same
decoder adapter.}
\label{tab:supp-kablation}
\end{table}

\section{Choice of foundation model}
\label{sec:supp-fm}
We test whether the mapper architecture is tied to UNI2-h by retraining it in
the CONCH, UNI-v1 and Virchow2 embedding spaces.
\Cref{tab:fmsweep} supports the mapper's portability across embedding spaces.
No space leads both slides: CONCH leads FID and SSIM on NCBI875, while
Virchow2 leads FID and UNI-v1 leads SSIM on NCBI884. UNI2-h remains within
$2.98$ FID and $0.010$ SSIM of the best arm on each slide. We therefore retain
UNI2-h because it is PixCell's native conditioning space, not because it wins
every metric. Qualitative examples are shown in \cref{fig:supp-gallery-fm}.

\begin{table}[tb]
\centering\small
\begin{tabular}{lcccc}
\toprule
& \multicolumn{2}{c}{NCBI875} & \multicolumn{2}{c}{NCBI884} \\
\cmidrule(lr){2-3}\cmidrule(lr){4-5}
FM & FID & SSIM & FID & SSIM \\
\midrule
Virchow2 & 101.58 & 0.318 & \textbf{44.16} & 0.207 \\
UNI-v1   & 111.19 & 0.310 & 51.12 & \textbf{0.213} \\
CONCH    & \textbf{86.67} & \textbf{0.321} & 45.56 & 0.206 \\
UNI2-h   & 89.65 & 0.313 & 45.86 & 0.203 \\
\bottomrule
\end{tabular}
\caption{\textbf{Four embedding spaces} on the two held-out lung slides. Each
complete pipeline uses the same mapper design and protocol with its own output
width and decoder LoRA. The UNI2-h row is the \ourmethod row of
\cref{tab:supp-lung-crosscond}.}
\label{tab:fmsweep}
\end{table}

\section{Cross-study breast transfer}
\label{sec:supp-breast}
\label{sec:exp-crossstudy}

This is our most stringent transfer setting, combining cross-study transfer
with a fully held-out patient cohort. Within HEST-1k~\cite{hest}, training uses
three 10x Xenium breast samples (TENX94/95/99; three patients, $1.82$M cells),
while evaluation uses three NCBI/GEO samples (NCBI783/784/785; two patients,
$427$k cells) on their shared 280-gene panel. Evaluation slides are excluded
from mapper and adapter training.

Under this compound shift, image fidelity degrades substantially. Prediction
FID is $97.1$--$105.6$, compared with $28.3$--$40.8$ for the decoder oracle,
while $\unicos$ falls to $0.40$--$0.48$ (\cref{tab:crossstudy}). The large
gap to the oracle suggests that much of the error comes from the predicted
embedding rather than the decoder (\cref{sec:supp-lora}).

Despite this degradation, the mapper retains a stable expression signal:
gene-influence rankings remain highly consistent across the held-out slides
(\cref{sec:supp-enrichment}).
The training and held-out samples differ in study, patient cohort and tissue
source, and receptor status varies within the held-out cohort.

\section{Stability of gene influence rankings}
\label{sec:supp-enrichment}

\begin{table}[h]
\centering\small
\resizebox{\linewidth}{!}{
\begin{tabular}{llcccc}
\toprule
Setting & $K$ & Overlap & \% & Chance \% & $p$ \\
\midrule
Lung        & 50  & 43/50   & 86 & 15 & $8{\times}10^{-40}$ \\
(343 genes) & 100 & 90/100  & 90 & 29 & $8{\times}10^{-59}$ \\
Cross-study & 50  & 40--46/50   & 80--92 & 18 & $\le\!1.4{\times}10^{-29}$ \\
(280 genes) & 100 & 86--98/100  & 86--98 & 36 & $\le\!1.3{\times}10^{-41}$ \\
\bottomrule
\end{tabular}}
\caption{Agreement of gene-influence rankings across held-out slides.
``Chance \%'' is the expected overlap under independent rankings. Cross-study
values span the three pairs of NCBI783/784/785, with the largest $p$ reported.}
\label{tab:supp-enrichment}
\end{table}

We ask whether one mapper depends on a stable set of genes across unseen
slides. Following \geneflow~\cite{geneflow}, we rank genes by accumulating the
absolute gradient of $\lVert\hat e\rVert_2^2$ w.r.t. per-cell
expression over $300$ tiles per slide. We compare top-$K$ rankings from the
same mapper across two cross-condition lung or three cross-study breast holdouts, using a
hypergeometric test against the measured gene panel.

\Cref{tab:supp-enrichment} shows that the mapper relies on the same genes
across held-out slides in both settings. Lung shares $86\%$ of its top $50$ genes against $15\%$ expected by
chance; cross-study breast shares $80$--$92\%$ against $18\%$, despite its
larger image-metric gap. Agreement remains significant at top $100$ in both
settings.

The shared rankings include recognizable tissue-associated genes. Lung
includes immune and stromal markers such as HLA-DQB1,
MS4A1, MARCO, ACKR1, MFAP5, SFRP2 and DCN; breast includes adipocyte and
stromal markers such as ADIPOQ, LPL, LEPR, LYVE1, MMP11, LAMA2 and CXCL12.

\section{Controllable tissue synthesis: density and proportion}
\label{sec:supp-synthesis}
The main paper demonstrates control by cell type. Here we vary density and
cell-type proportion, testing whether generated morphology follows each
change while everything else is kept fixed.

\begin{figure*}[ht]
\centering
\includegraphics[width=1.0\linewidth]{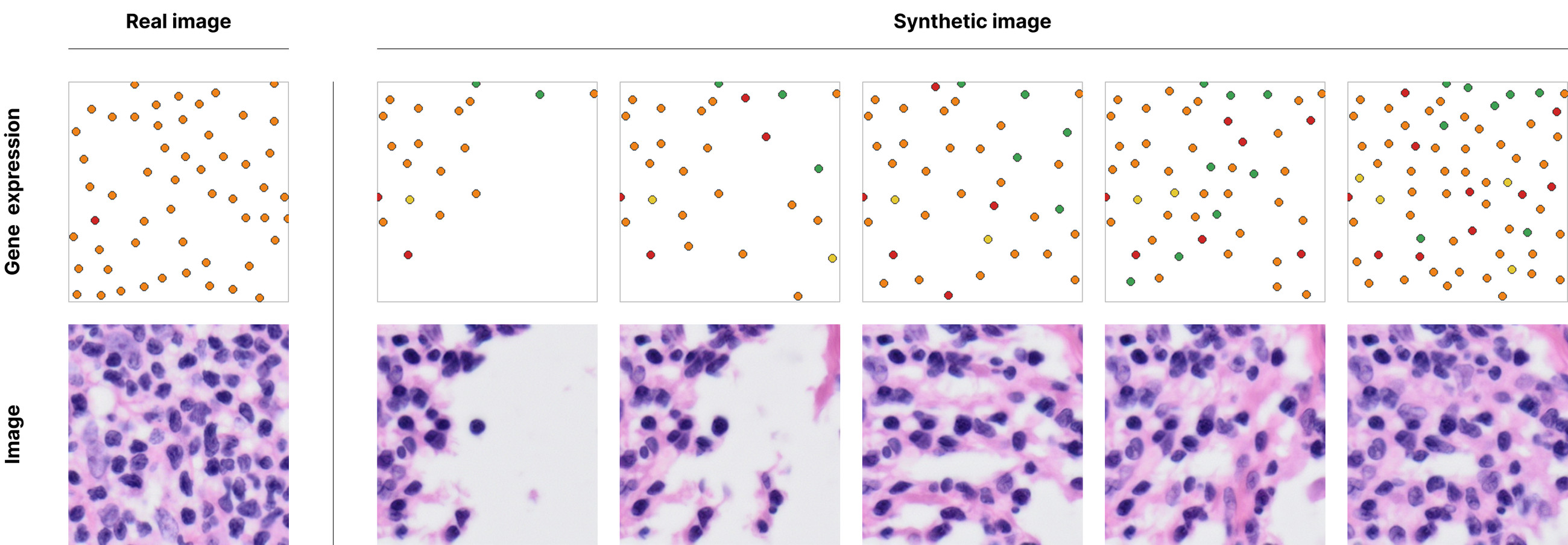}
\caption{\textbf{Cell density.} Column 1 is real tissue of this type at its
typical density and is an appearance reference. The remaining columns add
cells to a separate $19$-cell field while retaining its composition and
expression state. Each map appears above its generated image.}
\label{fig:synth-density}
\end{figure*}

\paragraph{Cell density.} We add cells to a real $19$-cell C2 field, sampling
from its T-cell-rich composition and expression state. The $19$--$66$ sweep
stays within this tissue's observed range (median $55$, reaching $81$).
Generated tissue tracks the requested density in
\cref{fig:synth-density}. CellViT~\cite{cellvit} nuclei increase
monotonically from $32$ to $79$ as the request rises from $19$ to $66$. We use
the same unretrained detector on real tiles matched by assay count.

\begin{table}[h]
\centering\small
\begin{tabular}{lccc}
\toprule
\multirow{2}{*}{Requested} & Generated & \multirow{2}{*}{Real tile} & Real      \\
& (CellViT) & & (CellViT) \\
\midrule
19 & 32 & 19 & 14 \\
28 & 38 & 24 & 25 \\
38 & 61 & 38 & 57 \\
48 & 67 & 48 & 61 \\
66 & 79 & 66 & 72 \\
\bottomrule
\end{tabular}
\caption{Density response under the same nucleus detector. Real tiles are
nearest-matched by Xenium assay count. Generated nuclei track the request
($r=0.965$), and detected real nuclei track assay counts ($r=0.942$).}
\label{tab:supp-density-cellvit}
\end{table}

\paragraph{Cell-type proportion.} Holding cell positions and the total count
fixed at $36$, we vary the fraction of cells given melanoma expression
against T cell expression.

\begin{figure*}[ht]
\centering
\includegraphics[width=1.0\linewidth]{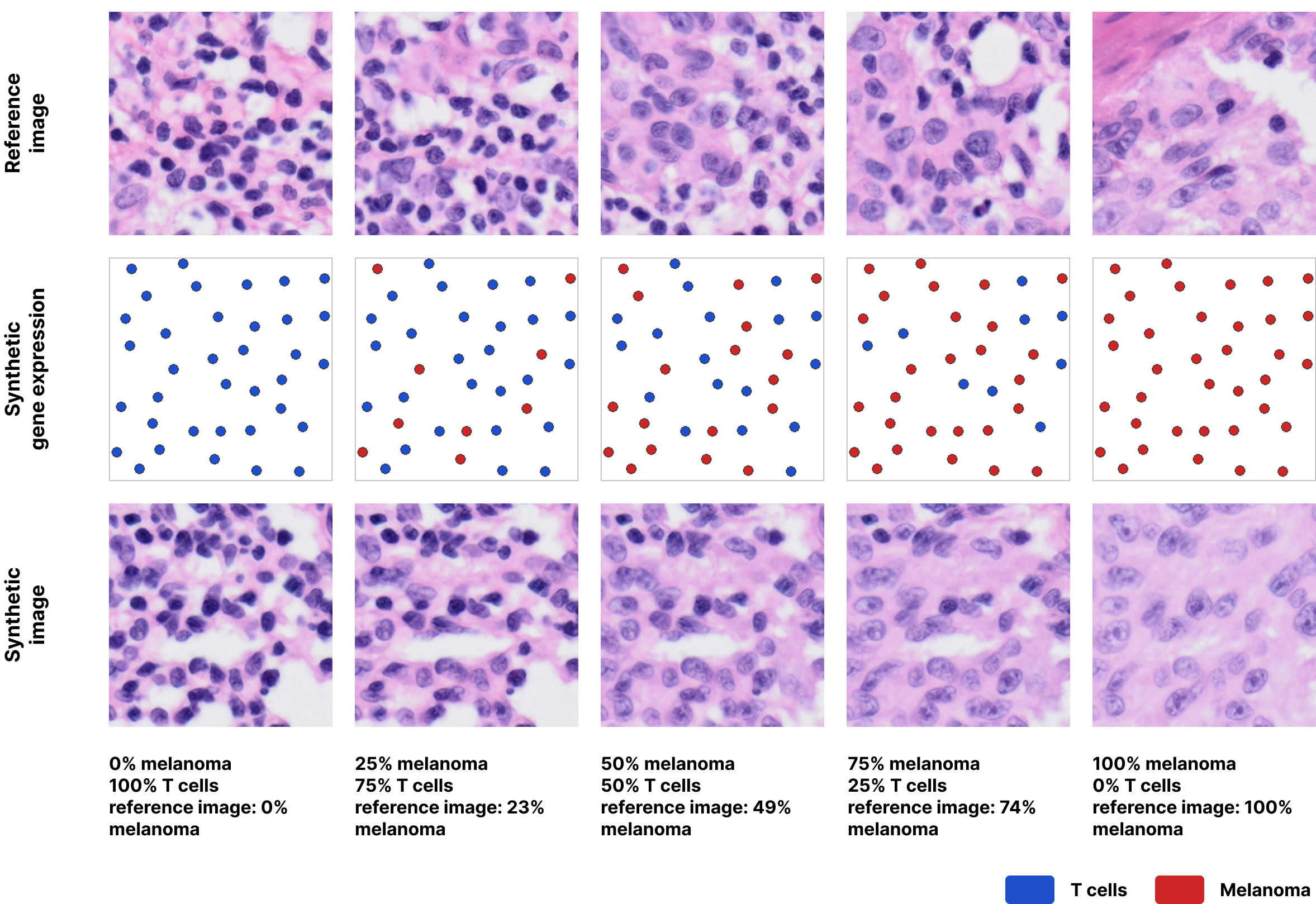}
\caption{\textbf{Cell-type proportion.} Positions and total count are fixed;
only the fraction of cells assigned melanoma rather than T-cell expression
changes. The top row gives an unpaired real reference nearest each requested
composition; the middle row is the synthetic expression input and the bottom
row its generated image.}
\label{fig:synth-composition}
\end{figure*}

In \cref{fig:synth-composition}, increasing melanoma fraction makes generated
nuclei larger, paler and more vesicular and the surrounding material more
eosinophilic. Real tiles show the same difference between the two extremes.
Together, the sweeps demonstrate that generated morphology responds
systematically to controlled changes in cell density and cell-type proportion.

\section{Qualitative generalization galleries}
\label{sec:supp-galleries}
\begin{figure*}[htp]
\centering
\includegraphics[width=\textwidth]{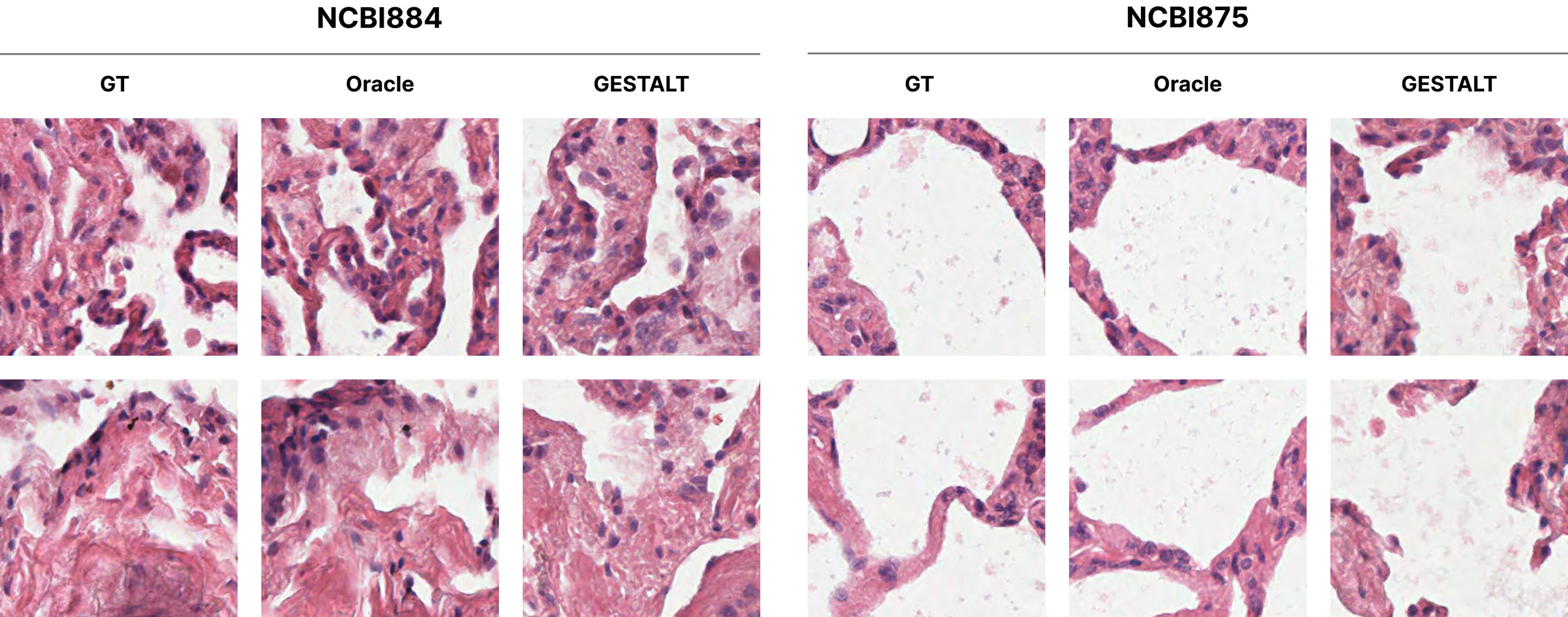}
\caption{\textbf{Lung cross-condition holdouts.}
Columns show ground truth (GT), oracle decoding from the real FM embedding
$D(e)$, and the expression-conditioned GESTALT prediction $D(\hat e)$.
Both slides are healthy controls from unseen patients, held out against a
training pool of $15$ diseased slides. Predictions retain broad tissue
organization and large open spaces but not exact local structure. Transfer is
uneven: the
full-slide prediction-to-oracle FID gap is $15.8$ on NCBI884 and $42.6$ on
NCBI875 (\cref{tab:supp-lung-crosscond}).}
\label{fig:supp-gallery-lung}
\end{figure*}

\begin{figure*}[htp]
\centering
\includegraphics[width=\textwidth]{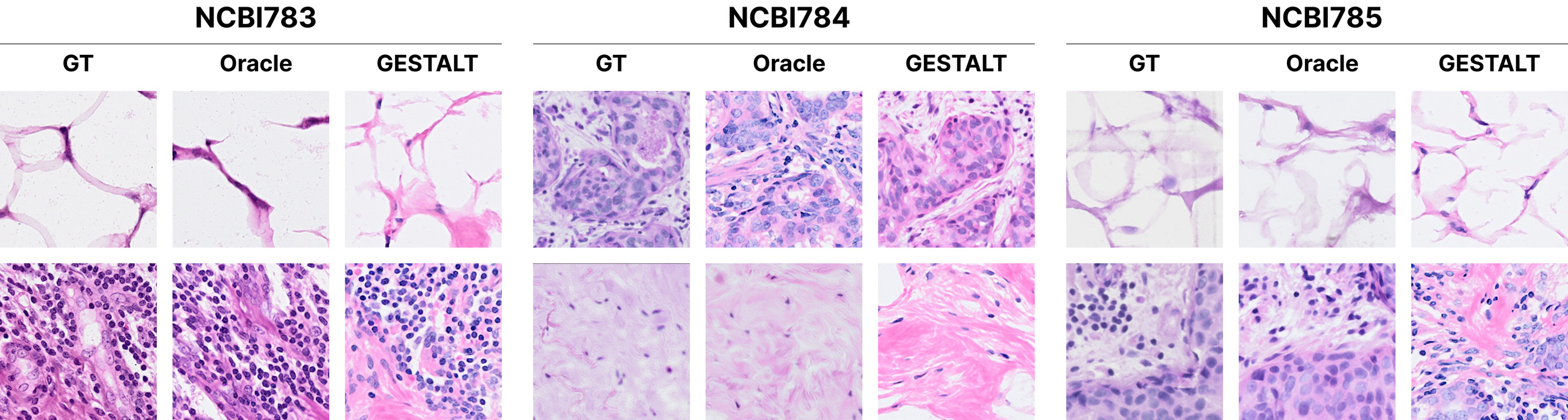}
\caption{\textbf{Cross-study transfer}, 10x
Xenium breast to GEO breast on the shared $280$-gene panel. Columns show
ground truth (GT), oracle decoding from the real FM embedding $D(e)$, and the
expression-conditioned GESTALT prediction $D(\hat e)$. The most visible error
in GESTALT predictions is a clear overall color shift from the ground truth;
oracle images more closely retain its color and coarse appearance, matching
the large prediction-to-oracle gap in \cref{tab:crossstudy}.}
\label{fig:supp-gallery-crossstudy}
\end{figure*}

\begin{figure*}[htp]
\centering
\includegraphics[width=1.0\textwidth]{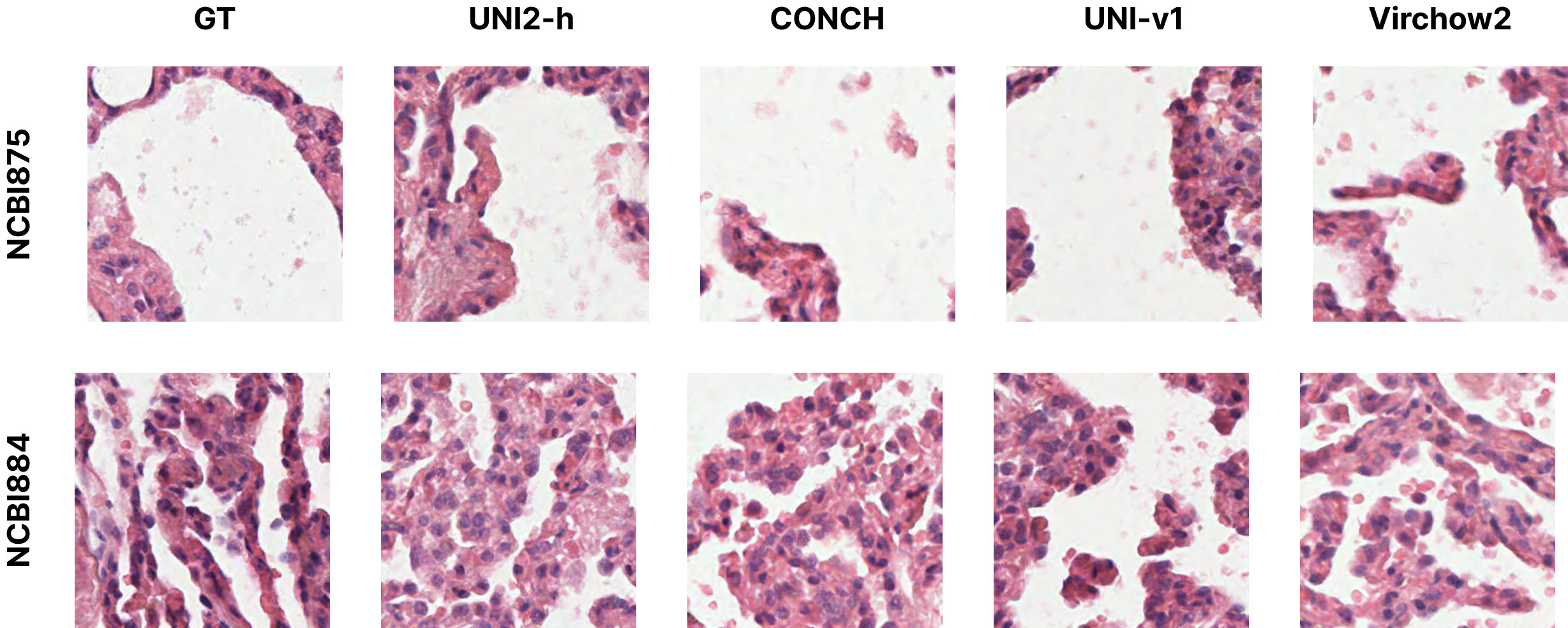}
\caption{\textbf{Four embedding spaces} on identical
held-out tiles of NCBI875 and NCBI884, each rendered by its own adapted
decoder. The same mapper design produces coarse tissue patterns in all four
spaces by changing only its output width, although local layout varies
(\cref{tab:fmsweep}).}
\label{fig:supp-gallery-fm}
\end{figure*}

These galleries complement the full-slide quantitative results for lung
transfer, cross-study breast transfer and foundation-model choice. Examples
were selected from seeded random candidate subsets and are illustrative; all
reported metrics use the complete evaluation sets. Comparisons always use the
same tiles. The three settings appear in
\cref{fig:supp-gallery-lung,fig:supp-gallery-crossstudy,fig:supp-gallery-fm}.
The lung and breast galleries place each ground-truth tile beside the oracle
$D(e)$ and expression-conditioned prediction $D(\hat e)$, following the
decomposition defined in \cref{sec:supp-lora}.

\begin{table*}[ht]
    \centering\small
    \setlength{\tabcolsep}{5pt}
    \begin{tabular}{lcccccccc}
    \toprule
    & & \multicolumn{4}{c}{\ourmethod} & \multicolumn{3}{c}{\emph{Oracle}} \\
    \cmidrule(lr){3-6}\cmidrule(lr){7-9}
    Test slide & Patient & FID $\downarrow$ & SSIM $\uparrow$ & FeatDist $\downarrow$ & \unicos $\uparrow$ & FID $\downarrow$ & SSIM $\uparrow$ & FeatDist $\downarrow$ \\
    \midrule
    NCBI783 & 5 & 97.06  & 0.313 & 17.54 & 0.479 & 28.28 & 0.287 & 13.80 \\
    NCBI784 & 4 & 103.35 & 0.318 & 16.08 & 0.408 & 40.83 & 0.363 & 14.23 \\
    NCBI785 & 4 & 105.59 & 0.376 & 16.36 & 0.403 & 28.75 & 0.428 & 13.19 \\
    \bottomrule
    \end{tabular}
    \caption{\textbf{Cross-study generalization}, 10x $\to$ GEO breast on the
    shared 280-gene panel, every held-out tile evaluated. Reported per slide
    because NCBI784 and NCBI785 are two sections of one patient.}
    \label{tab:crossstudy}
\end{table*}

\end{document}